\documentclass[letterpaper]{article} 
\usepackage{aaai2026}  
\usepackage{times}  
\usepackage{helvet}  
\usepackage{courier}  
\usepackage[hyphens]{url}  
\usepackage{graphicx} 
\usepackage{natbib}  
\usepackage{caption} 
\usepackage{algorithm}
\usepackage{algorithmic}

\usepackage{newfloat}
\usepackage{listings}
\DeclareCaptionStyle{ruled}{labelfont=normalfont,labelsep=colon,strut=off} 
\floatstyle{ruled}
\newfloat{listing}{tb}{lst}{}
\floatname{listing}{Listing}
\usepackage{amsmath,amssymb}
\usepackage{mathtools}
\usepackage{amsthm}
\usepackage{multirow}
\usepackage{xspace} 
\usepackage{subcaption}
\usepackage{xcolor}
\usepackage{tikz}
\usetikzlibrary{patterns}
\usetikzlibrary{external}
\usepackage[dvipsnames]{xcolor}
\usepackage{pgfplotstable, pgfmath} 
\usepackage{pgfplots} 
\pgfplotsset{compat=1.18} 
\newcommand{\dynotears}{\textsc{DyNoTears}\xspace}

\newcommand{\pcmci}{$\textsc{PcMci}^+$\xspace}
\newcommand{\varlingam}{\textsc{VARLiNGAM}\xspace}

\newcommand{\dbcms}{\textsc{DBCM}\xspace}

\newcommand{\nSHD}{\textsc{nSHD}\xspace}

\newcommand{\fone}{\textsc{F}1\xspace}
\newcommand{\aupr}{\textsc{Auprc}\xspace}
\newcommand{\erdos}{Erdős–Rényi\xspace}

\newcommand\ourmethod{\textsc{CaDyT}~}

\newcommand\tgraph{\mathcal{G}}

\newcommand\variables{\ensuremath{\mathbf{X}}}

\newcommand\observations{\ensuremath{\mathbf{Y}}}

\newcommand\trajectory{\ensuremath{\mathcal{T}}}
\newcommand\pa{\ensuremath{Pa}}
\newcommand\muff{\ensuremath{\mu}\text{-faithful}}
\newcommand\mum{\ensuremath{\mu}\text{-Markovian}}

\newcommand{\Models}{\ensuremath{\mathcal{M}}}

\newcommand{\lag}{\ensuremath{s}}
\newcommand{\gain}{\ensuremath{\Gamma}}

\newcommand{\Gright}{\ensuremath{M_{\oplus ij}}}

\newcommand{\dyn}{\textsc{Dyn}}

\DeclareMathOperator*{\argmin}{argmin}
\newcommand\timetrain{{[t_1:t_N]}}

\newtheorem{definition}{Definition}
\newtheorem{theorem}{Theorem}
\newtheorem{corollary}[theorem]{Corollary}
\newtheorem{assumption}{Assumption}
\newtheorem{lemma}[theorem]{Lemma}

\newcommand\independent{\protect\mathpalette{\protect\independenT}{\perp}}
\def\independenT#1#2{\mathrel{\rlap{$#1#2$}\mkern2mu{#1#2}}}
\usepackage{enumitem}

\title{Causal Structure Learning for Dynamical Systems with Theoretical Score Analysis}
\author{
    Nicholas Tagliapietra\textsuperscript{\rm{1,2}}, 
    Katharina Ensinger\textsuperscript{\rm{1}}, 
    Christoph Zimmer\textsuperscript{\rm{3}}, 
    Osman Mian\textsuperscript{\rm{4}}
}
\affiliations{
    \textsuperscript{\rm 1}Bosch Center for Artificial Intelligence, Renningen, Germany\\
    \textsuperscript{\rm 2}Computer Science Department, TU Darmstadt, Germany\\
    \textsuperscript{\rm 3}Baden-Wuerttemberg Cooperative State University Mannheim, Germany\\
    \textsuperscript{\rm 4}Institute for AI in medicine IKIM, Germany\\
    tagliapietra.nicholas@gmail.com
}

\begin{document}

\maketitle

\begin{abstract}
Real world systems evolve in continuous-time according to their underlying causal relationships, yet their dynamics are often unknown. Existing approaches to learning such dynamics typically either discretize time ---leading to poor performance on irregularly sampled data--- or ignore the underlying causality. We propose \ourmethod, a novel method for causal discovery on dynamical systems addressing both these challenges. In contrast to state-of-the-art causal discovery methods that model the problem using discrete-time Dynamic Bayesian networks, our formulation is grounded in Difference-based causal models, which allow milder assumptions for modeling the continuous nature of the system. \ourmethod leverages exact Gaussian Process inference for modeling the continuous-time dynamics which is more aligned with the underlying dynamical process. We propose a practical instantiation that identifies the causal structure via a greedy search guided by the Algorithmic Markov Condition and Minimum Description Length principle. Our experiments show that \ourmethod outperforms state-of-the-art methods on both regularly and irregularly-sampled data, discovering causal networks closer to the true underlying dynamics.
\end{abstract}

\section{Introduction} \label{sec:introduction}

\begin{figure}[t]
    \centering
    \includegraphics[width=0.9\linewidth]{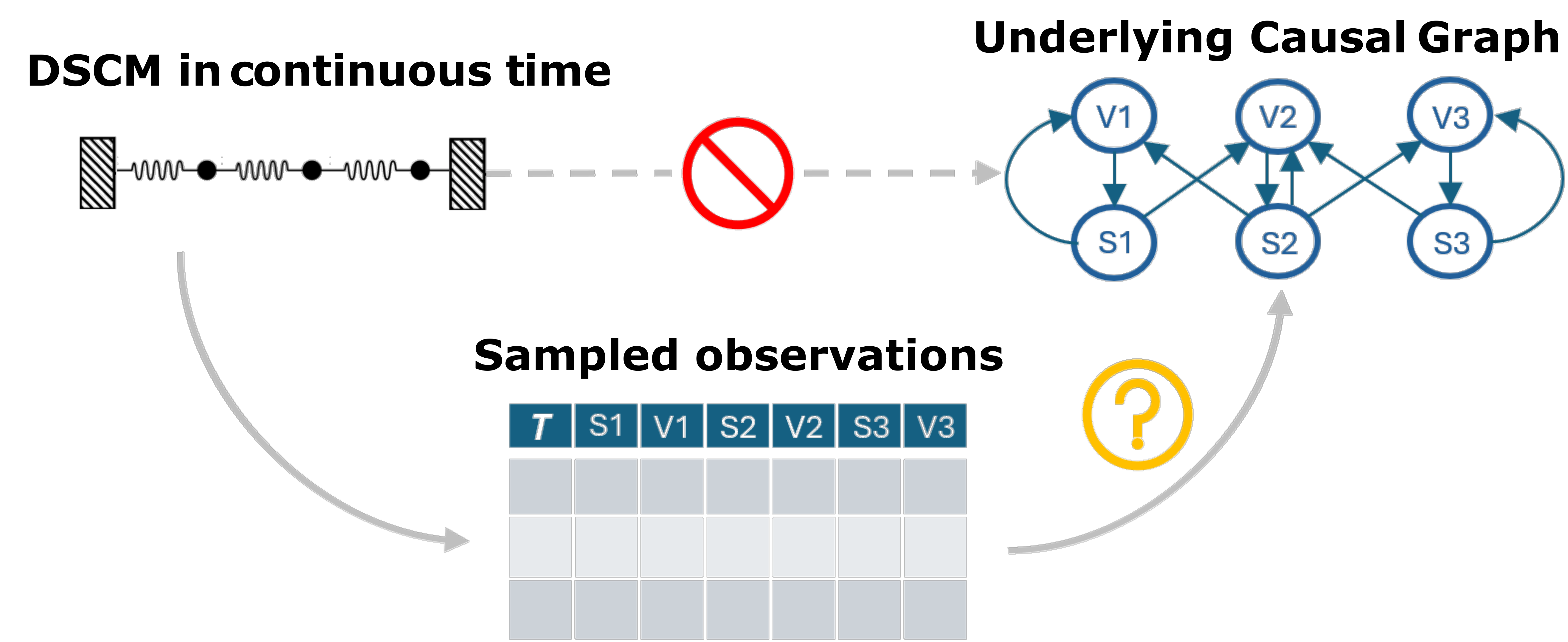}
    \caption{\ourmethod\!\! discovers the unknown causal structure (Top-right) using trajectories sampled from a continuous-time dynamical system (e.g.\, n-mass spring system). Our learned continuous-time model adapts to arbitrary timelines, including irregularly sampled ones.
    }
    \label{fig:motivation}
\end{figure}

Real-world physical systems are fundamentally governed by continuous-time dynamics \citep{strogatz:2000} with intrinsic causal mechanisms. For instance, in a mass-spring system as shown in Figure n:\ref{fig:motivation}, the position of each mass ($S_i$) induces a force influencing its own velocity ($V_i$) and the velocities of other masses connected to it with a spring. The position of each mass, however, depends only on its own velocity. 
This example remarks the importance of incorporating the directionality of causal relationships to achieve physical plausiblity in learned models. While differential equations are the de facto standard for modeling dynamical systems, inferring them from data remains challenging. Moreover, when leveraging data-driven approaches such as neural networks or Gaussian processes to learn these models, there are rarely guarantees that the true underlying dynamics will include causality, and spurious correlations might be incorporated by accident. These challenges highlight the need for a causal discovery framework for learning systems in continuous time.

Existing state-of-the-art for causal discovery on time-series focuses on learning the causal structure, but rarely captures the underlying continuous dynamics. Indeed, most methods learn a discretized version of the true dynamics~\cite{shimizu:10:varlingam,peters:13:timino,runge:20:pcmci,pamfil:20:dynotears} while assuming regular sampling, and are not designed for irregularly-sampled data. Dynamic-systems modeling, on the other hand, solve this issue, and current approaches adapt to irregularly-sampled data by learning a continuous-time model \citep{chen2019neuralordinarydifferentialequations, Hegde2022}. These methods, however, ignore causality and do not guarantee the generalization capabilities of causal models.

In this work, we tackle both challenges: we propose a novel approach capable of performing causal discovery on dynamical systems in a continuous-time fashion, relaxing the regular sampling assumption. We review the conditions under which dynamics can be modeled with Dynamic Structural Causal Models~\cite{mooij:13:odescm}, and build our method around those conditions. Our approach, \ourmethod\!\!, leverages the Gaussian Process-based framework developed by \citet{ensinger2024continuous} to learn a continuous-time model of the dynamics. We incorporate those in our novel score, that leverages the Algorithmic Markov Condition (AMC) postulate \cite{janzing:08:amc}. AMC allows us to identify the causes of a target variable as the ones providing the \emph{simplest} description, in terms of Kolmogorov complexity. Our score upper bounds the, otherwise uncomputable, Kolmogorov complexity via the Minimum Description Length principle~\cite{grunwald:07:mdl}. We then minimize this score via structure search for the underlying causal structure. Our contributions are as follows:
\begin{itemize}
    \item We propose a novel approach for causal discovery in continuous-time dynamical systems that addresses both irregular sampling and causal structure identification—two challenges previously tackled in isolation.
    \item We leverage the Gaussian Process framework for exact inference in continuous time, enabling nonparametric modeling of system dynamics. 
    \item We develop an end-to-end algorithm, \textbf{Ca}usal Discovery for \textbf{Dy}namic \textbf{T}imeseries (\ourmethod\!\!), combining Gaussian Process dynamical system modeling methods for continuous-time inference with structure search, enabling exact evaluation of dynamics while recovering true causal mechanisms. 
\end{itemize}

\section{Background}
In this section we introduce the basic formalism for dynamical systems described by differential equations, and how they are typically learned. Next, we provide a causal interpretation of dynamical systems through dynamic structural causal models (DSCM). We end this section by explaining how we can use the Algorithmic framework of~\citet{janzing:08:amc} to perform causal discovery over such dynamical systems defined using DSCMs.
\subsection{Dynamical Systems}\label{sec:dynamical_systems}
We consider a multivariate real-valued stochastic process $\variables^{(t)} = \{ X_1^{(t)}, \dots, X_D^{(t)} \}$ on a compact time interval, i.e.\, $t \in [0,T]$ and each $X_i^{(\cdot)} \in \mathbb{R}$. We characterize the evolution of the system's states through the framework of dynamical systems \citep{strogatz:2000}, where its dynamics are governed by a defined set of rules that control state transitions. In the continuous-time setting relevant to our work, the dynamics are typically formalized using systems of \textit{autonomous Ordinary Differential Equations} (ODE) of the form
\begin{equation}\label{eq:ode_system}
    \dot{\variables}^{(t)} = F(\variables^{(t)}) \quad \text{with }F: \mathbb{R}^D \rightarrow \mathbb{R}^D,
\end{equation}
where $\dot{\variables}^{(t)}$ is the time derivative of $\variables^{(t)}$ representing its rate of change at time $t$, and $F = (F_0, \dots, F_D)$ is the vector field describing the system's dynamics. A trajectory of a dynamical system is the path traced starting from an initial condition $\variables^{(0)}$ for which Eq.\eqref{eq:ode_system} holds. The system of ODEs induces causal dependencies between components of the dynamics (and/or trajectories) describing if and how each one influences another, called local dependency. \citet{bellot:22:neuraldag} provide the following formalization
\begin{definition}[Local dependency]
    Two components $X_i$ and $X_j$ are locally dependent given any other processes iff $X_i$ appears in the differential equation of $X_j$ i.e. $|\partial_i F_j| \neq 0$.
\end{definition}
Essentially, a component $X_j$ is locally dependent on component $X_i$ if $X_i$ directly influences $X_j$ in Eq.\eqref{eq:ode_system}, and independent otherwise. These dependencies entail an associated directed (potentially cyclic) graph, $\tgraph$, where the components $X_i \in \variables$  are the nodes and there is a directed edge from $X_i \rightarrow X_j$ if and only if $X_j$ is locally dependent on $X_i$.

\paragraph{Dynamics Model Learning:}  
In the field of Dynamics Model Learning, we aim at learning an approximation of the dynamics $F$ from discrete-time trajectory data. Since observations are discrete while the underlying dynamics are continuous, we must first discretize the ODE using a numerical integrator. This discretization enables matching continuous dynamics to discrete observations. In this work, we use multi-step integrators \cite{hairer2008solving} since they allow for exact GP inference.
They approximate a point in the trajectory $X^{(n+s)}$ by using a weighted sum of the last $s-1$ points $\bar{\variables}^{[n:n+s-1]} = \{\bar{\variables}^{(n)}, \dots, \bar{\variables}^{(n+s-1)}\}$ and are of the form
\begin{equation}\label{eq:numerical_integrator}
    \sum_{j=0}^{s}a_{jn}\bar{X}^{(n+j)} = \sum_{j=0}^{s}b_{jn}F(\bar{X}^{(n+j)}),
\end{equation}
where $a_{jn}$ and $b_{jn}$ are integrator-specific coefficients (e.g., Adams-Bashforth or Adams-Moulton methods).
The number of stages is determined by $s$ and often corresponds to the order of convergence \cite{hairer2008solving}.
 
\paragraph{Gaussian Processes} Gaussian Processes (GPs) \citep{10.5555/1162254} are a class of probabilistic models which describe a random function $F:\mathbb{R}^D \rightarrow \mathbb{R}$.  
They are a generalization of multivariate normal distributions, and analogously, are fully described by their mean function $m(x)$ and covariance function $k_{\theta}(x,y)$, where $\theta$ denotes the trainable parameters such as lengthscales. Further, the covariance function $k(\cdot, \cdot): \mathbb{R}^{D} \rightarrow \mathbb{R}^{D}$ can be a kernel such as the \textit{Radial Basis Function} (RBF) kernel, or the \textit{Polynomial} kernel. 
Here, we assume a zero-mean Gaussian process prior $F \sim \mathcal{N}(0, k_{\theta}(x,y))$. Further, a GP conditioned on a number of test points $\variables= \{\variables^{(1)}, \dots, \variables^{(N)}\}$ and their associated observations $\observations = \{Y^{(1)}, \dots, Y^{(N)}\}$ still follows (by definition) a multivariate Gaussian distribution, with mean $\mu(X^{*})$ and variance $\Sigma(X^{*})$ derived as
\begin{align}
     \mu(X^{*}) &= k(X^{*})^T(K + \lambda I)^{-1} Y,\label{eq:gp_regression_mean}\\
     \Sigma(X^{*}) & = k(X^{*}, X^{*}) - k(X^{*})^T(K + \lambda I)^{-1} k(X^{*}),\label{eq:gp_regression_var}
\end{align}
where $K \in \mathbb{R}^{N \times N}$ is the covariance matrix evaluated at points $X^{[1:N]}$ i.e. $K_{ij} = k(X^{(i)}, X^{(j)})$. This conditioning operation is called \textit{Gaussian Process Regression} (GPR).

When applied to Dynamics Model Learning, different variants of GPR are used for approximating the dynamics function $F$ in Eq.\ref{eq:ode_system} by conditioning the GP on measured trajectory points. 
Following \citet{ensinger2024continuous}, multi-step integrators \cite{hairer2008solving} enable exact GP inference and make it possible to evaluate the learned $F$ conditioned on observations. 
To do so, \citet{ensinger2024continuous} derived kernels leveraging multi-step integrators as, 
\begin{align}\label{eq:gp_kernels}\begin{split}
K(\bar{X}^{(n)}, \bar{X}^{(m)}) &=\mathbf{b}_n^\top\, k(\bar{X}^{[n:n+s]}, \bar{X}^{[m:m+s]})\, \mathbf{b}_m,\\
k(\bar{X}^{*}) & = \mathbf{b}_n^\top\, k(\bar{X}^{*}, \bar{X}^{[n:n+s]}),
\end{split}\end{align}
where $\mathbf{b}_n \in \mathbb{R}^s$ is the n-th column of the matrix $B \in \mathbb{R}^{s\times N}$ containing the integration coefficients, and $k(\bar{X}^{[i:j]}, \bar{X}^{[k:l]})$ is a block matrix obtained after evaluating the chosen kernel between $\bar{X}^{[i:j]}$ and $\bar{X}^{[k:l]}$. In essence, given $N$ (potentially irregularly sampled) trajectory points, $\variables^{[t_1,\dots,t_N]}$ 
we can evaluate the dynamics at any point $\bar{X}^{*}$ by performing the GPR in Eq.\eqref{eq:gp_regression_mean},\eqref{eq:gp_regression_var} using the kernels in Eq.~\eqref{eq:gp_kernels}
\begin{equation}\label{eq:gp_dynamics}
F_D(\variables^{\star}|\variables^{t_1:t_N})\sim \mathcal{N}(\mu_{post}(\variables^{\star}),\Sigma_{post}(\variables^{\star})),
\end{equation}
where the equations for mean and covariance depend on the chosen multi-step integrators. The advantage of using multi-step integrators within GPR is that, depending on their order, they permit a better representation of the continuous dynamics $F$ and allow for evaluations of $F$ at any test point $\variables^{*}$.  

Most dynamics model learning methods discussed above ignore the causal structure and fail to preserve local dependencies. This could lead to inaccurate predictions of a component $X_j$ when an independent, unrelated component $X_i$ undergoes a distribution shift. In the following we show how to equip these methods to take causal relations into account.

\subsection{Dynamic Structural Causal Models}
The system of ODEs shown in Eq.\eqref{eq:ode_system} induces different local dependencies (Def. 1) between stochastic processes. This dependency can be thought of as modularization of the system. The set of individual component trajectories is called modular if the admitted trajectories of the entire system can be detached into trajectories of individual components~\cite{mooij:13:odescm}. Under the dynamical stability assumption, which states that asymptotic dynamics of the system of ODE's converge to a unique element irrespective of the initial conditions, a system of ODE can be converted to a \textit{Dynamic Structural Causal Model} (DSCM)~\cite{rubenstein:16:dscm}, defined as follows.
\begin{definition}[\citealp{rubenstein:16:dscm}]\label{def:dscm}
    Let $\dyn_i$ be the trajectory for component $X_i$, and let $\dyn = \bigcup_{X_i \in \variables} \dyn_{i}$ be a modular set of trajectories. A deterministic Dynamic Structural Causal Model (DSCM) on time indexed variables \variables~taking values in \dyn~is a collection of equations.

    \begin{equation}\label{eq:dscm}
        \mathcal{S} : \{ X_i = F_i(\pa_i), ~~ X_i\in \variables \}~,
    \end{equation}
where $\pa_i \subseteq \variables \setminus \{X_i\}$ and each $F_i$ is a map that outputs the trajectory of the effect variable $X_i$ in terms of the trajectory of its direct causes $\pa_i$.
\end{definition}
In the above definition, the parents $\pa_i$ should be interpreted as \emph{direct causes} of $X_i$, and the function $F_i$ as \emph{causal mechanism} that maps the direct causes to the effect. Further, self-loops in the causal graph are allowed, although they do not explicitly appear in Def.\ref{def:dscm}, which describes instead the stationary asymptotic behavior of the system.  
In essence in a DSCM, we explicitly decompose  Eq.~\eqref{eq:ode_system} into multiple simpler sub-equations, one for each component, as a direct consequence of local consistency. \citet{rubenstein:16:dscm}
proves that it is possible to derive a DSCM that allows us to reason about the asymptotic dynamics of the underlying ODE if the dynamic stability assumption holds, and impossible otherwise. Hence going forward, we assume dynamic stability. We adopt the instantaneous gradient assumption, meaning that causal relationships are encoded in the time derivatives of $\variables$. This assumption is more aligned with the ODE structure and milder than the instantaneous-effect assumption, which requires instantaneous dependencies between variables. Consequently, our proposal is more aligned with Difference-based Causal Models (DBCM)~\cite{voortman:10:dbcm} as opposed to the Dynamic Bayesian Networks.

Let trajectory $\trajectory \in \mathbb{R}^{N \times D}$ be a sequence of observations of $\variables$ sampled from a DSCM over time-steps $\{t_1, ..., t_{N}\}$ and denoted by $\trajectory = \{\variables^{(1)}, ..., \variables^{(N)}\}$, we aim to find the underlying directed graph $\tgraph$ of process interactions entailed by the DSCM and learn the dynamics of each variable using only its causal parents. Doing so is impossible unless we make assumptions on how $\trajectory$ was generated~\cite{pearl:09:causality}. To that end, we assume causal sufficiency, i.e.\, for each component $X_i$ there are no unobserved components that influence $X_i$. In addition we assume $\tgraph$~is $\mum$~with respect to $\trajectory$ which implies that local independencies of the underlying DSCM are reflected in $\trajectory$. Conversly we assume $\trajectory$ is $\muff$~with respect to $\tgraph$ which implies that all local independencies that we find in $\trajectory$ also hold in $\tgraph$. Together these assumptions ensure that independence statements derived from $\trajectory$ can be interpreted as absence of edges in $\tgraph$, thereby letting us deduce local independencies~\cite{mogensen:20:usep}. Let $\phi_{max}$ be the highest frequency present among all the components within $\mathcal{S}$ and define $\Delta_{max} = \max(t_{i+1} - t_i)~\mathit{for}~1\leq\! i\! < N$, we assume that  $\Delta_{max} \in (0, \frac{1}{2\phi_{max}})$, meaning that \trajectory~has been sampled at a rate finer than its critical frequency. Next, we explain how to learn the underlying causal structure entailed by a DSCM.

\paragraph{Information Theoretic Causal Discovery}
Information theoretic causal discovery relies on the algorithmic Markov condition (AMC)~\citep{janzing:08:amc}, and is grounded in Kolmogorov complexity. The Kolmogorov complexity of a binary string $x$ is the length of the shortest binary program $p^*$ that outputs $x$ and halts on a universal Turing machine $U$ ~\cite{kolmogorov:65:kc, vitanyi:02:kstructurefunction}. For a probability distribution $P$, this complexity $K(P)$ is the length of the shortest program that outputs $P(x)$ to within precision $q$ on input $\langle x, q \rangle$. Formally,
\begin{equation*}
    K(P) = \min_{p \in \{0,1\}^*} \left\{ ||p|| : |U(p, x, q) - P(x)| \leq \frac{1}{q} \right\}
\end{equation*}
The AMC states that a graph $\tgraph$ over $\variables$, with joint distribution $P$, is an admissible causal graph only if the shortest description of $P$ factorizes as $ K(P(X_1, \ldots, X_D)) = \sum_{j=1}^{D} K(P(X_j \mid \pa_j)) + \mathcal{O}(1)$. Thus, the true causal graph minimizes Kolmogorov complexity i.e. each $\pa_j$ provides the tersest description of its causal child. 

While Kolmogorov complexity is not computable due to the halting problem, it can be bounded from above in a statistically well-founded way using the Minimum Description Length (MDL) principle~\citep{grunwald:04:MDLTutorial}. \citet{marx:21:formally} show that for sample sizes approaching infinity, the true causal graph can be identified by minimizing an appropriate lossless MDL-score. 

Given a model class $\mathcal{M}$, the MDL principle selects the optimal model $M \in \mathcal{M}$ for data $D$ by minimizing the total description length: $L(D, M) = L(M) + L(D \mid M),$ where $L(M)$ represents the number of bits required to describe the model $M$, and $L(D \mid M)$ is the number of bits needed to describe the data $D$ once the model $M$ is known. Existing methods have already used this algorithmic model to discover causal graphs for non-timeseries data with reasonable success~\cite{kaltenpoth:19:coca, mian:21:globe, mameche:23:linc}. 

\section{MDL for Dynamical Systems}\label{sec:theory}
To use information theoretic causal discovery with dynamical systems, we need to build a suitable MDL score for timeseries trajectories sampled from a DSCM. We will do so for the well known case of Additive Noise Models~\cite{hoyer:09:anm} and assume that we have 
\begin{equation}\label{eq:causalmodel}
        X_{i}^{(t)} = \hat{X}_{i}^{(t)} + \nu_{i}^{(t)} \;,
\end{equation}
with $\hat{X}_{i}^{(t)}$ following the dynamical system defined in Eq.~\eqref{eq:ode_system} for $t \in [0, T]$ and $\nu_{i}^{(t)} \sim \mathcal{N}(0, \sigma^2_i)$ being independent gaussian noise terms such that $\nu_{i} \independent X_i~\forall i$ and $\nu_{i} \independent \nu_{j}~\forall i,j$. This setup can be interpreted as a noisy observation model in dynamical systems, where a deterministic process is perturbed by independent additive noise. Given a trajectory \trajectory~of size $N$ for (possibly irregular) time steps $\{t_1,..., t_{N}\}$, we want to find the model  such that
\begin{align}
\label{eq:global-score0} M^* \!&=\argmin_{M \in \Models}~~L(\trajectory, M), \\
\label{eq:global-score1} \!&= \argmin_{M \in \Models}\left(\! L(M) + \sum_{i=1}^D L(X_i^\timetrain\! \mid\! \pa_i,F_i)\!\right)\!, \\
\label{eq:global-score2} \!&= \argmin_{M \in \Models}\left(\! L(M) + \sum_{i=1}^D L(\mathbf{\nu}_i )\! \right) \; ,
\end{align}
where we use the notation $X_i^{[a:b]}$ to denote the values for $X_i$ from time-steps $a$ to $b$ included. The summation term in Eq.\eqref{eq:global-score1} follows from Eq.\eqref{eq:dscm} and measures compression of each component given its causal parents and the parametrization enforced by $M$. We simplify this in Eq.\eqref{eq:global-score2} to show that encoding a component given the model reduces to encoding the noise terms. To be able to use MDL as a practical stand-in for Kolmogorov complexity inside AMC, we need to define a model class and an encoding scheme measuring the complexity of the class resp. data under that model class. This we do next.
\paragraph{Encoding the Model} 
The model cost $L(M)$ consists of a global cost $L_{global}(M)$ plus the sum of the local model costs for each individual variable $X_i$, i.e.\, $\sum_{i=1}^{D} L(M_i)$. For a pre-specified integrator of step-size $\lag$, the global cost measures the complexity of storing the initial $\lag$ samples of a given trajectory $\trajectory$. Formally,
\begin{equation}\label{eq:global_model_cost}
    L_{global}(M) = \log N + r_d \cdot D \cdot \lag \;, 
\end{equation}
where we encode the integrator stepsize $\lag$ using $\log N$ bits and first $\lag$ samples of the trajectory. We assign a fixed cost of $r_d$ bits to each component value $X_i^{(t)} \in \trajectory^{[t_1:t_\lag]}$.

For each $X_i$, we define a local model $M_i$ where we store its causal parents and the parameters of the structural equation $F_i$ in Eq.~\eqref{eq:dscm}.  We encode $M_i$ as 
\begin{equation}
\label{eq:model-node}
L(M_i) = L_{\mathbb{N}}(||\pa_i||) + ||\pa_i||\log D + L_{F}(F_i) \; , 
\end{equation}
where $L_\mathbb{N}$ encodes the number of parents using the MDL-optimal encoding for integers $z \geq 0$~\citep{rissanen:83:integers}. It is defined as $L_\mathbb{N}(z) = \log^* z + \log c_0$, where $\log^* z = \log z + \log \log z + \ldots$ and only positive terms are considered. Further, $c_0$ is a normalization constant to ensure the Kraft-inequality holds~\citep{kraft:49:device}. Next, we identify those $||\pa_i||$ variables and encode the function $F_i$ over them.

\textbf{Encoding the Functions} To compute each local model $M_i$, we learn a continuous dynamics function $F_i$ for each $X_i$ by using the GPR scheme developed by \citet{ensinger2024continuous}. We do so due to two main reasons namely 1) they offer a natural fit for modeling systems of ODEs by learning a continuous model and not a discrete one and are therefore capable of handling irregularly sampled trajectories and, 2) their non-parametric modeling-nature saves us from imposing parametric assumptions on $F_i$. Each model $M_i$ regresses the dynamics related to variable $X_i$ from its causal parents $\pa_i$. In essence, we learn a GP as defined in Eq.\eqref{eq:gp_dynamics} using the kernels in Eq.\eqref{eq:gp_kernels}, yielding an estimator in the form 
\begin{equation}
F_i(X^{t_\star}_i|\pa_i^{[t_1:t_\star)})\sim \mathcal{N}\big(\mu_{post}(X^{t_\star}_i),\Sigma_{post}(X^{t_\star}_i)\big),
\end{equation}
where we can estimate the dynamics of a $X_i$ at an arbitrary point $t_\star$ using the history of $\pa_i^{[t_1:t_\star)}$ up to this point. Once a GP is trained, $X_i^{(t)}$ can be estimated by numerically integrating the dynamics $F_i(\cdot)$ learned by the GP over an arbitrary timeline. We formally score $F_i(\cdot)$ as,
\begin{equation}\label{eq:LF}
    L_F(F_i)\! =\! \log \left(\frac{1}{r_\lambda}\right)\! \frac{N(N-1)}{2}\! +\! L_\phi([\boldsymbol{\alpha_i},\boldsymbol{\beta_i},\boldsymbol{\Lambda_i}]) \;. 
\end{equation}
The components of $F_i$ comprise the kernel matrix $K_i$ and the corresponding length-scale and noise-variance parameters $\boldsymbol{\alpha_i}$ and $\boldsymbol{\beta_i}$, computed from $\pa_i$. Since the integrator coefficients are deterministically obtained from $\boldsymbol{\alpha_i}$, $\boldsymbol{\beta_i}$, $K_i$, and the initial trajectory in Eq.~\eqref{eq:global_model_cost}, they need not be stored. To store $K_i$ efficiently, we apply Singular Value Decomposition $K_i = V_i \Lambda_i V_i^\top$, where $V_i$ and $\Lambda_i$ denote the orthonormal eigenvector and diagonal eigenvalue matrices, respectively. The orthonormal matrix $V_i \in \mathbb{R}^{N \times N}$ can be represented using at most $N(N-1)/2$ rotation angles at a predefined precision $r_\lambda$. Finally, the length scales $\boldsymbol{\alpha_i}$, variances $\boldsymbol{\beta_i}$, and eigenvalues in $\boldsymbol{\Lambda_i}$ are encoded using $L_p$.

\textbf{Encoding the Parameters} To encode the length-scale resp. eigenvalues obtained from the SVD, we use the score proposed by~\citet{marx:19:slope} for encoding parameters up to a user-specified precision $p$. We have
\begin{equation}\label{eq:encode-params}
L_{p}(\theta) = 2||\theta|| + \sum_{i = 1}^{||\theta||}  L_{\mathbb{N}}(|\rho_i|) + L_{\mathbb{N}}(\lceil \theta_i \cdot 10^{\rho_i} \rceil) \; ,
\end{equation}
with $\rho_i$ being the smallest integer such that $|\theta_i| \cdot 10^{\rho_i} \ge 10^p$. To simplify, $p = 2$ implies that we consider first two digits of the parameter. We need two bits to store the signs of $\rho_i$ and the parameter, then we encode the shift $\rho_i$ and the shifted parameter $\theta_i$.
\paragraph{Encoding Data Given Model}
As a final step in constructing a lossless score, we encode the residual noise that remains after our the model has encoded the underlying causal structure and data-generating process. As our goal is to minimize the variance of the residuals across the timeseries trajectory, we encode each $\nu_i$ as zero mean Gaussian noise using the encoding provided by~\citet{grunwald:04:MDLTutorial}, formally
\begin{equation}\label{eq:residuals}
L(\nu_{i}) = \frac{N}{2} \left( \frac{1}{\ln 2} + \log 2 \pi \hat{\sigma}_{i}^2  \right) \;,
\end{equation}
where $\hat{\sigma}_{i}^2$ is the empirical estimate of residual noise variance. Combining the above, we obtain a lossless MDL score for a timeseries trajectory modeled using a causal graph. 

\subsection{Theoretical Analysis}
While lack of computability of Kolmogorov complexity impedes us from directly providing identifiability guarantees using the Algorithmic Markov Condition (AMC), we can still independently prove that our score acts as a valid regularized log-likelihood score with an upperbound asymptotically similar to the BIC score~\cite{schwarz:78:bic}. Doing so however necessitates two additional assumptions.
\begin{assumption}[Finite dimensions]\label{assm:finite-support}
$K$ is finite-dimensional.
\end{assumption}
\begin{assumption}[Bounded hyperparameters and precision]\label{assm:boundedprecision}
The length scale parameters $\alpha_i$, the variance parameters $\beta_i$ are upper bounded. All precisions $|\rho_i|$ are upper bounded.
\end{assumption}

Kernel classes satisfying Assm.~(\ref{assm:finite-support}) include, Polynomial kernels, Wendland kernels~\cite{wendland:95:kernel}, Buhmann kernels~\cite{buhmann:03:kernel}, Truncated resp. Random Fourier kernels~\cite{rahimi:07:rfkernel}, and Nyström Approximated kernels~\cite{williams:00:nystrom-kernel}. Intuitively, Assms.~\ref{assm:finite-support} and~\ref{assm:boundedprecision} ensure that the cost of storing the eigenvalues in Eq.~\eqref{eq:LF} scales sub-linearly with the number of trajectory points, which is necessary to provide theoretical guarantees. Let $C = \log \left(\frac{1}{r_\lambda}\right) \cdot \frac{N(N-1)}{2} +  2||\theta||  + \frac{N}{2}\left( \frac{1}{\ln2} + \log(2\pi)\right)$, 
we show the following.
\begin{lemma}
   \label{lemm:mec}
   Given a DSCM $\mathcal{S}$, let $\trajectory$ be a trajectory generated from $\mathcal{S}$ and let $\bar{L}(\trajectory,M_j) = L(M_j) + L(\nu_{j}) - C$. If Assms.~\ref{assm:finite-support},\ref{assm:boundedprecision} hold, it holds asymptotically
    \begin{equation*}
     \bar{L}(\trajectory,M_j) \leq c_0 \cdot N \cdot log(\hat{\sigma}_j^2) +   c_1^{(j)}  log(N) + c_2^{(j)}.
    \end{equation*}
    with constants $c_0,c^{(j)}_1, c_2^{(j)}$ independent of $N$. 
\end{lemma}

\begin{theorem}
    \label{th:mec}
    Given a DSCM $\mathcal{S}$, let $\trajectory$ be a trajectory generated from $\mathcal{S}$ and let $\bar{L}(\trajectory,M) =  \sum_{i=1}^{D} \bar{L}(\trajectory,M_i)$. If Assms.~\ref{assm:finite-support},\ref{assm:boundedprecision} hold, it holds asymptotically.
    \begin{equation*}
     \bar{L}(\trajectory,M)  \leq  c_0 \cdot N \cdot log(\hat{\sigma}^2) +   c_1  log(N) +  c_2.
    \end{equation*}
    with constants $c_0,c_1$, and $c_2$ independent of $N$ and $\bar{L}(\trajectory,M)$ asymptotically is a valid regularized log-likelihood score according to Definition A.2. 
\end{theorem}
\begin{corollary}
    \label{corr:main}
    Model selection using ${L}(\trajectory,M)$ is equivalent to model selection using $\bar{L}(\trajectory,M)$.
\end{corollary}
We provide the proof in Appendix \ref{appendix:theoretical_results}. 
Intuitively, our proof shows that the upper bound of $L(\trajectory,M)$ behaves like sum of component-wise regularized log-likelihood scores (such as the BIC). These guarantees, however, only hold if we score all graphs over $\trajectory$. This is an intractable brute-force approach because the search space grows super-exponentially in the number of variables. To nevertheless have a practical instantiation for minimizing $L(\trajectory,M)$, we use a general greedy structure search algorithm which we describe next.
\section{The \ourmethod Algorithm}
We present the score-based method \textbf{Ca}usal Discovery for \textbf{Dy}namic \textbf{T}imeseries (\ourmethod\!\!) for discovering causal graphs of multivariate continuous-valued dynamical systems. We incorporate our proposed score into a common three-step search procedure~\citep{mian:21:globe,mameche:23:linc} namely, edge scoring, forward and backward search. This is the next best alternative to exhaustive search for our case. The well known Greedy Equivalence Search~\cite{chickering:02:ges} is not built for timeseries and methods for topological search~\cite{wang:igsp,xu:25:topic} do not directly apply to cyclic systems. We provide full pseudocode in Appendix \ref{sec:pseudocode}.

We start with the edge-ranking phase that computes the gain for all pairwise causal connections. The gain $\gain_{ij}$, between each pair $X_i$ and $X_j$, is given by
\begin{equation}
    \gain_{ij} = L(\trajectory,M) - L(\trajectory,\Gright)  \label{eq:delta1} \; ,
\end{equation}
where \Gright~ implies model $M$ with edge $X_i \to X_j$ included. Intuitively, the higher the $\gain_{ij}$, the more confident we are that this is the correct causal edge. The edge scoring phase returns a priority queue of tuples $(\gain_{ij},(X_i,X_j))$ ordered by decreasing gain.

The forward search iteratively adds the highest-scoring edge from the priority queue. After adding an edge \(X_i \to X_j\), all incoming edges to \(X_j\) are re-evaluated using Eq.~\eqref{eq:delta1} and updated in the queue. Before inclusion, each edge is tested for statistical significance using the no-hypercompression inequality~\citep{grunwald:07:mdl}. The search terminates when no further additions improve the score. The subsequent backward phase prunes redundant edges by removing any whose deletion increases \(L(\trajectory, M)\), continuing until no such edges remain. We then return the final graph. 

\ourmethod has overall computational complexity of $\mathcal{O}(N^3D^3\log D)$ where $N^3$ derives from our choice of non-parametric regression functions (GPs) for $L_F$. In Appendix \ref{appendix:complexity}
we give a detailed derivation on the computational complexity and show how it is at least on-par with existing methods. \ourmethod, moreover, can be inherently parallelized, and we implement it as such, resulting in a fast runtime. 
\section{Related Work} \label{sec:related_work}
\paragraph{GPs for time-series modeling}
Most works in this field address discrete-time dynamics~\cite{Pilco2011,Wang2005} while
we aim to learn GP dynamics models in continuous time. However, there is also work that addresses continuous-time modeling~\cite{Heinonen2018,Hegde2022, ridderbusch2023}. \citet{glass2024safeactivelearninggaussian} apply the inference scheme \cite{Hegde2022} to active learning.
We leverage the method proposed by \citet{ensinger2024continuous} due to the beneficial properties of exact inference even under irregular sampling. None of the mentioned approaches, as opposed to our work, can discover the underlying causality in dynamical systems.
\begin{figure}[t!]
    \includegraphics[width=0.95\linewidth]{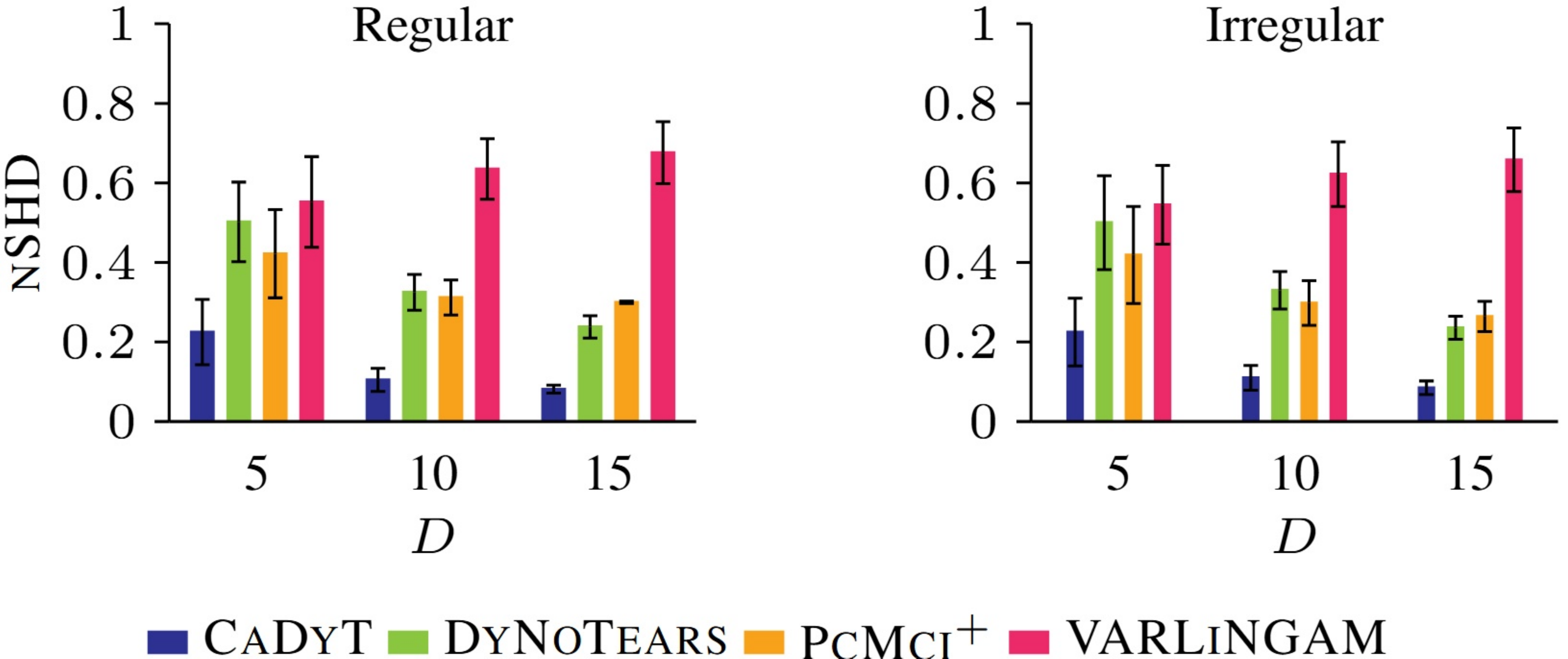}
    \caption{[Lower is better, \nSHD] for random graphs of sizes $D \in \{5,10,15 \}$ for regularly sampled data (left) and irregularly sampled data (right). \ourmethod finds graphs closer to the ground truth resulting in a lower \nSHD.}
    \label{fig:NHD}
\end{figure}%
\paragraph{Causal models for time-series } Causal discovery from time-series data has received active research interest over the past decade. Early methods focused on Granger-causality (G-causality)~\cite{granger:69:causality}, which implies that $X_i$ G-causes $X_j$ if including past of $X_i$ helps in predicting the present of $X_j$. Methods have been designed for both linear~\cite{geweke:82:mvgc,barrett:10:mvgc} and non-linear~\cite{nauta:19:tcdf} causal models. Methods that are not explicitly based on G-causality lie in the category of constraint-based methods~\cite{chu:08:anltsm,sun:15:ocse,runge:20:pcmci}, score-based methods~\cite{pamfil:20:dynotears} or noise-based methods~\cite{shimizu:10:varlingam,peters:13:timino}. These methods assume regularly sampled discrete trajectories and overlook the continuous-time structure of time series and the conditions required to reliably learn the underlying dynamics.

\citet{voortman:10:dbcm} were among the first ones to study under which conditions dynamical systems allow for a causal interpretation, and proposed Difference-based causal models (\dbcms). \dbcms differ from well known dynamic Bayesian networks in that they force all causation to go through derivatives. This idea was extended by \citet{mooij:13:odescm} to show that (D)SCMs can model the asymptotic behavior of systems of ODEs with limitations on the possible interventions under \dbcms. Those limitations are addressed in recent works \cite{blom:20:ccm,cinquini_2025}.

Our proposal aims to model a DBCM rather than a dynamic Bayesian network. In contrast to the existing work, our work is uniquely positioned at the intersection of dynamical system learning and causal discovery as it provides a theory-backed approach to modeling the former, while allowing for learning the laws of the process via the latter.

\begin{figure}[t!]
	 \includegraphics[width=0.95\linewidth]{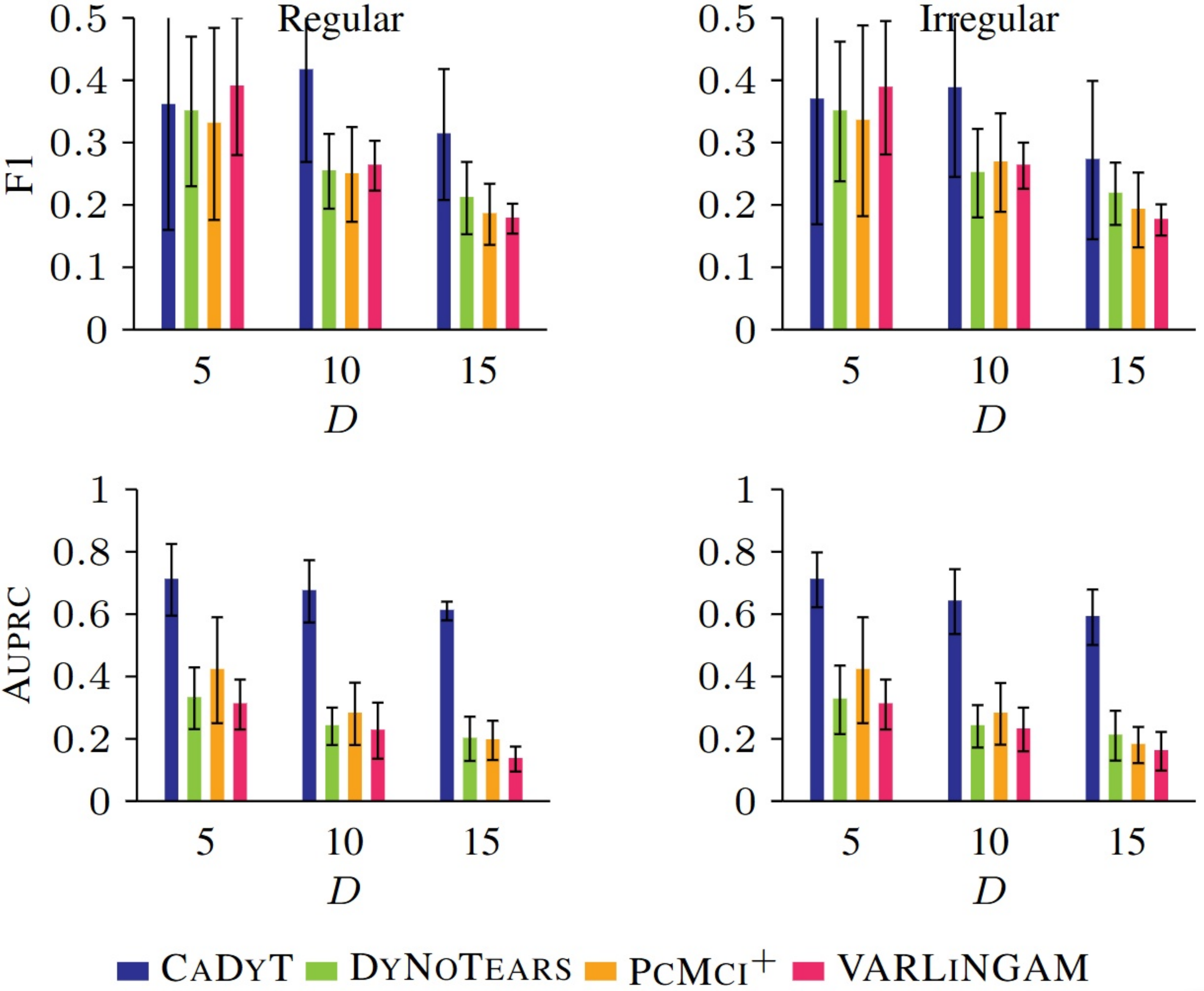}
    \caption{[Higher is better, \fone score (top) and \aupr (bottom)] for random graphs of sizes $D \in \{5,10,15 \}$ for regularly (left) resp. irregularly sampled data (right). \ourmethod improves over baselines in terms of \fone. \ourmethod having high \aupr indicates higher confidence about the correct causal edges as opposed to spurious ones.}
    \label{fig:f1auprc}
\end{figure}%

\section{Experiments} \label{sec:experiments}
\paragraph{Setup} We instantiate \ourmethod using parallelized greedy search. We perform our experiments with GPs with RBF-Kernel leveraging explicit Adams-Bashforth (AB) integrators of order \, $s \in \{1,2,3\}$. Even though Thm.~1 applies to finite-dimensional Kernels, our choice is motivated by RBF-Kernel's exact-inference capabilities. Even with over-regularization that could result due to the use of RBF-Kernel, we find that we still outperform the competition. We include the results for \ourmethod using Polynomial kernels in the supplementary material. We compare \ourmethod with a variety of baselines: The constraint-based \pcmci~\cite{runge:20:pcmci} using non-parametric Kernelized independence test, the score-based \dynotears~\cite{pamfil:20:dynotears}, and the noise-based \varlingam~\cite{shimizu:10:varlingam}. 

We generate synthetic data using Diamond structure (4 variables) and Erdős–Rényi random graphs with and without cycles for $D \in \{5, 10, 15\}$ for both regular and irregular timelines. To evaluate the predicted structures we measure the \textit{Structural Hamming Distance} (\textsc{SHD})~\citep{Tsamardinos2006TheMH} which counts the edge mismatch in true and predicted structures. For comparability across structures of different sizes, we normalize \textsc{SHD} between $0$ and $1$ by dividing with $D^2$ and call this \nSHD. To evaluate precision and recall over predicted edges we use the \fone metric, and use  \textit{Area Under Precision Recall Curve} (\aupr) to assess how correct each method is on the edges it is most confident about. We repeat all experiments 20 times with different seeds and report the mean. We report results for the more challenging setting of \erdos graphs with both regular and irregular sampling in the manuscript and postpone full experimental details to Appendix \ref{appendix:timelines}.

\paragraph{Results} \label{sec:results} We perform a sanity check and assess the robustness of \ourmethod to false-positives. We generated $10$ graphs made of $4$ independent ODEs. \ourmethod with AB3 never discovered a single spurious edge. Surprisingly, baselines report causal edges for independent data. \varlingam found spurious edges 30\% of times, whereas \pcmci and \dynotears 60\% and 100\% respectively. 

\begin{figure}[t]
	\includegraphics[width=0.9\linewidth]{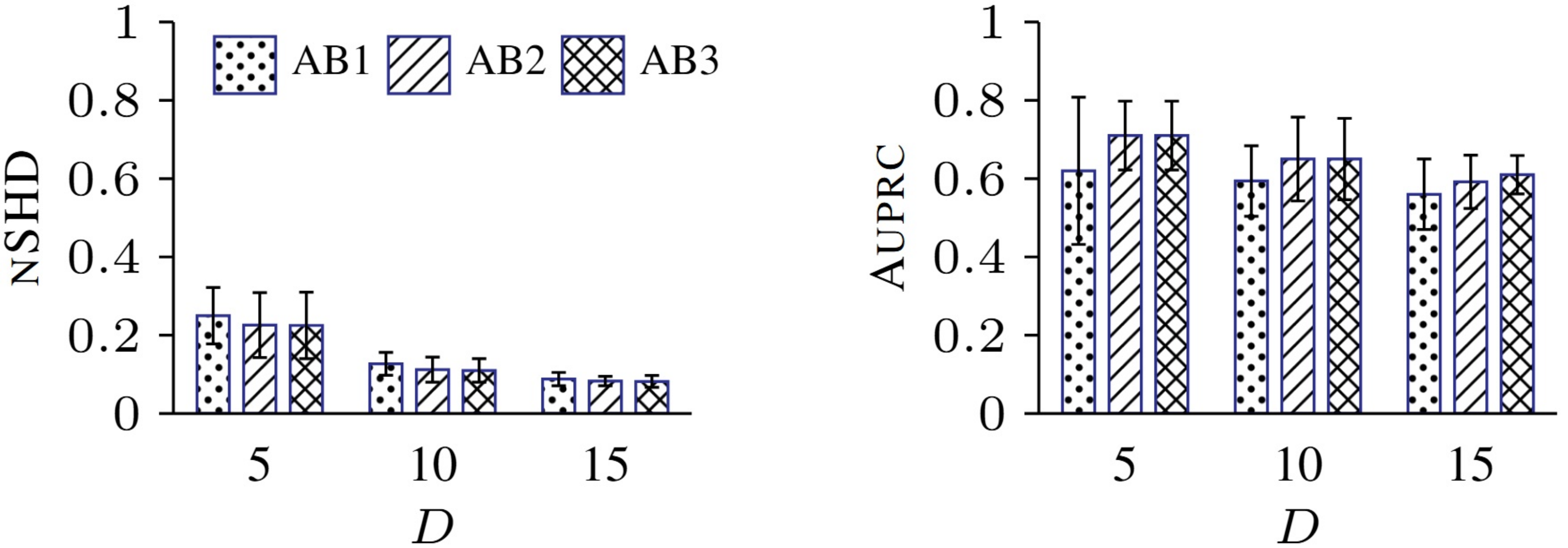}
	\caption{[\nSHD (left) Lower is better, \aupr (right) Higher  is better] for Adams-Bashforth integrator of order $s \in \{1,2,3\}$ for irregularly sampled data. \ourmethod shows gradual improvement with higher order integrators.}
    
	\label{fig:integrators}
\end{figure}%
Next we sample random dynamical systems and evaluate how well the methods can discover the underlying causal structure. We sample cyclic and acyclic dynamical systems with equal probability and report the results. We report \nSHD in Figure n:~\ref{fig:NHD} where we observe that \ourmethod consistently outperforms all baselines by a clear margin. Both \dynotears and \pcmci demonstrate a high false positive rate which worsens the \nSHD. Overall, all methods worsen slightly on irregularly-sampled data, but \ourmethod still remains best by a visible margin.

To evaluate how well we perform in-terms of precision and recall, we report the \fone score in Figure n:~\ref{fig:f1auprc}. We see that \ourmethod is on-par with the baselines for variable sizes $5$, and continues to be robust to false-positives by maintaining a high precision as network size increases. Competing methods on the other hand have very low precision as they tend to predict spurious edges quite frequently.

While \nSHD and \fone score could give us a summarized picture of how well the methods perform, we are also interested in how correct are the methods on the causal relationships that they are most confident about. To that end, we calculate the \aupr metric by ordering the predicted edges of each algorithm in decreasing confidence. For \ourmethod this confidence is computed using Eq.~\eqref{eq:delta1}, For \pcmci we use the p-values associated with each edge whereas for \dynotears resp. \varlingam we use the strength of the causal edge as present in the predicted adjacency matrix. Looking at the results in Figure n:~\ref{fig:f1auprc} we see that \ourmethod again performs well across benchmarks. A high \aupr indicates that \ourmethod is mostly correct about high-confidence edges.

\paragraph{Effect of Integration Order} To study the effect of the integration scheme, we conduct an ablation study whose result we report in Figure n:\ref{fig:integrators}. We compare integrators on irregular timelines and see that higher-order integrators (AB2/AB3) generally outperform AB1. This is consistent with the domain knowledge that higher-order integration schemes approximate the underlying continuous-time dynamics better.

We observed that addition of cycles into the data-generating process affects the methods differently. We find \ourmethod tends to improve with the order of integration, where versions leveraging higher-order schemes (AB2 and AB3) stay robust or even improve, whereas the lower-order AB1 variant deteriorates. 

\paragraph{Oscillators and Chaotic Systems} We further test on the simulated $2$-mass spring system, its analogous real-world counterpart \citep{schmidt2009freeform}, and the hyper-chaotic Rössler Oscillator. To stay fair to the baselines, we use regular sampling. We report results averaged over $10$ runs in Table n:~\ref{tab:nonlinear} which shows that \ourmethod outperforms the competing methods and finds causal structures closer to the underlying dynamics (lower SHD), while denoting high confidence about true edges (high \aupr).
\begin{table}[t!]
\centering
\caption{[Higher is better, \aupr] for the simulated 2-mass spring system (DblMass), for the real double-linear system (DblLinear), and for the Rössler Oscillator (Rössler).}
\label{tab:nonlinear}
\begin{tabular}{cccc}
\hline
\textbf{Method} &  DblMass & DblLinear  & Rössler \\ \hline
\dynotears        & 0.22       & 0.59 & 0.34      \\
\pcmci            & 0.24       & 0.30 & 0.22 \\
\varlingam        & 0.39       & 0.44 & 0.30 \\ 
\ourmethod (ours) &  \textbf{0.79} & \textbf{0.79} & \textbf{0.55} \\ \hline
\end{tabular}
\end{table}
\section{Discussion and Conclusions} \label{sec:conclusions}
We proposed \ourmethod, a method for uncovering causal structure in continuous-time dynamics from discrete trajectory data. Our approach combines the exact inference framework of~\citet{ensinger2024continuous} with the Algorithmic Markov Condition~\citep{janzing:08:amc}. We proved that our score is a valid regularized log-likelihood score (Def.A.2)
with an upper-bound asymptotically similar to the BIC, and demonstrated empirically that our score outperforms existing methods.  Going forward we see potential lines of improvements as future work.

First, we used greedy graph search and the Adams-Bashforth integrators as the two main components for \ourmethod\!\!. We do not, however, claim that these to be optimal choices. It is possible that alternate search strategies resp. integrators yield better performance and we aim to investigate such alternative approaches. Second, while using the proposal of \citet{ensinger2024continuous} allowed us to naturally work with both regular and irregular-sampled trajectories, it has been known to be sensitive to high amount of noise. We conjecture that we could further improve the results by evolving the machinery to be robust to noise. Third, we could extend \ourmethod to relax the causal sufficiency assumption by searching for instantaneous edges such that both edge directions give a high score, post backward search. This could potentially point to a hidden confounder. 
Last, currently \ourmethod assumes ODEs can be well-modeled by a GP regression under noisy observation model. A recently proposed extension of DSCMS to chaotic models~\cite{boeken2024dynamicstructuralcausalmodels} could allow us to relax this assumption further and is a potentially rewarding line of future work.

\section*{Acknowledgments}
Nicholas Tagliapietra and Katharina Ensinger are supported by Robert Bosch GmbH. Osman Mian is supported by the German Federal Ministry of Research, Technology and Space (DECIPHER-M, 01KD2420C).
\bibliography{aaai2026}

\begin{thebibliography}{67}
\providecommand{\natexlab}[1]{#1}

\bibitem[{Barrett, Barnett, and Seth(2010)}]{barrett:10:mvgc}
Barrett, A.~B.; Barnett, L.; and Seth, A.~K. 2010.
\newblock Multivariate Granger causality and generalized variance.
\newblock \emph{Physical Review E—Statistical, Nonlinear, and Soft Matter Physics}, 81(4).

\bibitem[{Bellot, Branson, and van~der Schaar(2022)}]{bellot:22:neuraldag}
Bellot, A.; Branson, K.; and van~der Schaar, M. 2022.
\newblock Neural graphical modelling in continuous-time: consistency guarantees and algorithms.
\newblock arXiv:2105.02522.

\bibitem[{Blom, Bongers, and Mooij(2020)}]{blom:20:ccm}
Blom, T.; Bongers, S.; and Mooij, J.~M. 2020.
\newblock Beyond Structural Causal Models: Causal Constraints Models.
\newblock In Adams, R.~P.; and Gogate, V., eds., \emph{Proceedings of The 35th Uncertainty in Artificial Intelligence Conference}, volume 115 of \emph{PMLR}, 585--594. PMLR.

\bibitem[{Boeken and Mooij(2024)}]{boeken2024dynamicstructuralcausalmodels}
Boeken, P.; and Mooij, J.~M. 2024.
\newblock Dynamic Structural Causal Models.
\newblock arXiv:2406.01161.

\bibitem[{Brouillard et~al.(2020)Brouillard, Lachapelle, Lacoste, Lacoste-Julien, and Drouin}]{brouillard:20:dcdi}
Brouillard, P.; Lachapelle, S.; Lacoste, A.; Lacoste-Julien, S.; and Drouin, A. 2020.
\newblock Differentiable causal discovery from interventional data.
\newblock \emph{Advances in Neural Information Processing Systems}, 33.

\bibitem[{Cavanaugh and Neath(1999)}]{cavanaugh:99:sic}
Cavanaugh, J.~E.; and Neath, A.~A. 1999.
\newblock Generalizing the derivation of the Schwarz information criterion.
\newblock \emph{Communications in Statistics-Theory and Methods}, 28(1).

\bibitem[{Chen et~al.(2019)Chen, Rubanova, Bettencourt, and Duvenaud}]{chen2019neuralordinarydifferentialequations}
Chen, R. T.~Q.; Rubanova, Y.; Bettencourt, J.; and Duvenaud, D. 2019.
\newblock Neural Ordinary Differential Equations.
\newblock arXiv:1806.07366.

\bibitem[{Chickering(2002)}]{chickering:02:ges}
Chickering, D.~M. 2002.
\newblock Optimal structure identification with greedy search.
\newblock \emph{JMLR}, 3.

\bibitem[{Chu, Glymour, and Ridgeway(2008)}]{chu:08:anltsm}
Chu, T.; Glymour, C.; and Ridgeway, G. 2008.
\newblock Search for Additive Nonlinear Time Series Causal Models.
\newblock \emph{Journal of Machine Learning Research}, 9(5).

\bibitem[{Cinquini et~al.(2025)Cinquini, Beretta, Ruggieri, and Valera}]{cinquini_2025}
Cinquini, M.; Beretta, I.; Ruggieri, S.; and Valera, I. 2025.
\newblock A Practical Approach to Causal Inference over Time.
\newblock \emph{Proceedings of the AAAI Conference on Artificial Intelligence}, 39: 14832–14839.

\bibitem[{Claeskens and Hjort(2008)}]{claeskens:08:book}
Claeskens, G.; and Hjort, N.~L. 2008.
\newblock Model selection and model averaging.
\newblock \emph{Cambridge books}.

\bibitem[{Deisenroth and Rasmussen(2011)}]{Pilco2011}
Deisenroth, M.~P.; and Rasmussen, C.~E. 2011.
\newblock PILCO: a model-based and data-efficient approach to policy search.
\newblock In \emph{Proceedings of the 28th ICML}, ICML'11. Omnipress.

\bibitem[{Ensinger et~al.(2024)Ensinger, Tagliapietra, Ziesche, and Trimpe}]{ensinger2024continuous}
Ensinger, K.; Tagliapietra, N.; Ziesche, S.; and Trimpe, S. 2024.
\newblock Exact inference for continuous-time Gaussian process dynamics.
\newblock In \emph{Proceedings of the Thirty-Eighth AAAI Conference on Artificial Intelligence and Thirty-Sixth Conference on Innovative Applications of Artificial Intelligence and Fourteenth Symposium on Educational Advances in Artificial Intelligence}, AAAI'24/IAAI'24/EAAI'24. AAAI Press.

\bibitem[{Geweke(1982)}]{geweke:82:mvgc}
Geweke, J. 1982.
\newblock Measurement of linear dependence and feedback between multiple time series.
\newblock \emph{Journal of the American statistical association}, 77(378).

\bibitem[{Glass, Ensinger, and Zimmer(2024)}]{glass2024safeactivelearninggaussian}
Glass, L.; Ensinger, K.; and Zimmer, C. 2024.
\newblock Safe Active Learning for Gaussian Differential Equations.
\newblock arXiv:2412.09053.

\bibitem[{Granger(1969)}]{granger:69:causality}
Granger, C. W.~J. 1969.
\newblock Investigating Causal Relations by Econometric Models and Cross-spectral Methods.
\newblock \emph{Econometrica}, 37(3): 424--438.

\bibitem[{Grunwald(2004)}]{grunwald:04:MDLTutorial}
Grunwald, P. 2004.
\newblock A tutorial introduction to the minimum description length principle.
\newblock arXiv:math/0406077.

\bibitem[{Gr{\"u}nwald(2007)}]{grunwald:07:mdl}
Gr{\"u}nwald, P.~D. 2007.
\newblock \emph{The minimum description length principle}.
\newblock MIT press.

\bibitem[{Hairer, N{\o}rsett, and Wanner(2008)}]{hairer2008solving}
Hairer, E.; N{\o}rsett, S.; and Wanner, G. 2008.
\newblock \emph{Solving Ordinary Differential Equations I: Nonstiff Problems}.
\newblock Springer Series in Computational Mathematics. Springer Berlin Heidelberg.
\newblock ISBN 9783540566700.

\bibitem[{Haughton(1988)}]{haughton:88:curveexpBIC}
Haughton, D.~M. 1988.
\newblock On the choice of a model to fit data from an exponential family.
\newblock \emph{The annals of statistics}.

\bibitem[{Hedge et~al.(2022)Hedge, Yildiz, L{\"a}hdesm{\"a}ki, Kaski, and Heinonen}]{Hegde2022}
Hedge, P.; Yildiz, C.; L{\"a}hdesm{\"a}ki, H.; Kaski, S.; and Heinonen, M. 2022.
\newblock Variational multiple shooting for Bayesian ODEs with Gaussian processes.
\newblock In \emph{Proceedings of the 38th Conference on Uncertainty in Artificial Intelligence (UAI 2022), PMLR}, Proceedings of Machine Learning Research, 790--799. United States: JMLR.
\newblock Conference on Uncertainty in Artificial Intelligence, UAI ; Conference date: 01-08-2022 Through 05-08-2022.

\bibitem[{Heinonen et~al.(2018)Heinonen, Yildiz, Mannerstr{\"o}m, Intosalmi, and L{\"a}hdesm{\"a}ki}]{Heinonen2018}
Heinonen, M.; Yildiz, C.; Mannerstr{\"o}m, H.; Intosalmi, J.; and L{\"a}hdesm{\"a}ki, H. 2018.
\newblock Learning unknown ODE models with Gaussian processes.
\newblock In \emph{Proceedings of the 35th ICML, ICML 2018}, volume~5 of \emph{PMLR}, 3120--3132. United States: International Machine Learning Society.

\bibitem[{Hoyer et~al.(2009)Hoyer, Janzing, Mooij, Peters, and Sch\"{o}lkopf}]{hoyer:09:anm}
Hoyer, P.; Janzing, D.; Mooij, J.~M.; Peters, J.; and Sch\"{o}lkopf, B. 2009.
\newblock Nonlinear causal discovery with additive noise models.
\newblock In \emph{NeurIPS}, volume~21. Curran.

\bibitem[{Huang et~al.(2018)Huang, Zhang, Lin, Sch{\"o}lkopf, and Glymour}]{huang:18:gges}
Huang, B.; Zhang, K.; Lin, Y.; Sch{\"o}lkopf, B.; and Glymour, C. 2018.
\newblock Generalized score functions for causal discovery.
\newblock In \emph{ACM SIGKDD}.

\bibitem[{Hyv{\"a}rinen et~al.(2010)Hyv{\"a}rinen, Zhang, Shimizu, and Hoyer}]{shimizu:10:varlingam}
Hyv{\"a}rinen, A.; Zhang, K.; Shimizu, S.; and Hoyer, P.~O. 2010.
\newblock Estimation of a structural vector autoregression model using non-Gaussianity.
\newblock \emph{Journal of Machine Learning Research}, 11(5).

\bibitem[{Janzing and Sch\"{o}lkopf(2010)}]{janzing:08:amc}
Janzing, D.; and Sch\"{o}lkopf, B. 2010.
\newblock Causal inference using the Algorithmic Markov Condition.
\newblock \emph{IEEETPAMI}, 56: 5168--5194.

\bibitem[{Kakade, Seeger, and Foster(2005)}]{kakade:05:gp-bounds}
Kakade, S.~M.; Seeger, M.~W.; and Foster, D.~P. 2005.
\newblock Worst-Case Bounds for Gaussian Process Models.
\newblock In Weiss, Y.; Sch\"{o}lkopf, B.; and Platt, J., eds., \emph{NeurIPS}, volume~18. MIT Press.

\bibitem[{Kaltenpoth and Vreeken(2019)}]{kaltenpoth:19:coca}
Kaltenpoth, D.; and Vreeken, J. 2019.
\newblock We Are Not Your Real Parents: Telling Causal from Confounded by MDL.
\newblock In \emph{SDM}. SIAM.

\bibitem[{Kashyap(1982)}]{kashyap:82:optimal}
Kashyap, R.~L. 1982.
\newblock Optimal choice of AR and MA parts in autoregressive moving average models.
\newblock \emph{IEEE Transactions on Pattern Analysis and Machine Intelligence}, 2.

\bibitem[{Kass and Raftery(1995)}]{kass:95:BICAnalysis}
Kass, R.~E.; and Raftery, A.~E. 1995.
\newblock Bayes factors.
\newblock \emph{Journal of the american statistical association}, 90(430).

\bibitem[{Kolmogorov(1965)}]{kolmogorov:65:kc}
Kolmogorov, A.~N. 1965.
\newblock Three approaches to the quantitative definition ofinformation’.
\newblock \emph{Problems of information transmission}, 1(1).

\bibitem[{Kraft(1949)}]{kraft:49:device}
Kraft, L.~G. 1949.
\newblock \emph{A device for quantizing, grouping, and coding amplitude-modulated pulses}.
\newblock Ph.D. thesis, Massachusetts Institute of Technology.

\bibitem[{Lu, Zhang, and Yuan(2021)}]{lu:21:optimalsearch}
Lu, N.~Y.; Zhang, K.; and Yuan, C. 2021.
\newblock Improving causal discovery by optimal bayesian network learning.
\newblock In \emph{Proceedings of the AAAI Conference on Artificial Intelligence}, volume~35.

\bibitem[{Mameche, Kaltenpoth, and Vreeken(2023)}]{mameche:23:linc}
Mameche, S.; Kaltenpoth, D.; and Vreeken, J. 2023.
\newblock Learning Causal Models under Independent Changes.
\newblock In Oh, A.; Naumann, T.; Globerson, A.; Saenko, K.; Hardt, M.; and Levine, S., eds., \emph{Advances in Neural Information Processing Systems}, volume~36, 75595--75622. Curran Associates, Inc.

\bibitem[{Martin, Buhmann, and Ablowitz(2003)}]{buhmann:03:kernel}
Martin, B.; Buhmann, M.; and Ablowitz, J. 2003.
\newblock Radial basis functions: theory and implementations.
\newblock \emph{Cambridge University (ISBN: 0-521-63338-9.)}.

\bibitem[{Marx and Vreeken(2019{\natexlab{a}})}]{marx:19:sloppy}
Marx, A.; and Vreeken, J. 2019{\natexlab{a}}.
\newblock Identifiability of Cause and Effect using Regularized Regression.
\newblock In \emph{KDD}. ACM.

\bibitem[{Marx and Vreeken(2019{\natexlab{b}})}]{marx:19:slope}
Marx, A.; and Vreeken, J. 2019{\natexlab{b}}.
\newblock Telling cause from effect by local and global regression.
\newblock \emph{KAIS}, 60(3): 1277--1305.

\bibitem[{Marx and Vreeken(2021)}]{marx:21:formally}
Marx, A.; and Vreeken, J. 2021.
\newblock Formally Justifying {MDL}-based Inference of Cause and Effect.
\newblock arXiv:2105.01902.

\bibitem[{Mian, Marx, and Vreeken(2021)}]{mian:21:globe}
Mian, O.; Marx, A.; and Vreeken, J. 2021.
\newblock Discovering Fully Oriented Causal Networks.
\newblock In \emph{AAAI}.

\bibitem[{Mogensen and Hansen(2020)}]{mogensen:20:usep}
Mogensen, S.~W.; and Hansen, N.~R. 2020.
\newblock Markov equivalence of marginalized local independence graphs.
\newblock \emph{The Annals of Statistics}, 48(1).

\bibitem[{Mooij, Janzing, and Sch{\"o}lkopf(2013)}]{mooij:13:odescm}
Mooij, J.~M.; Janzing, D.; and Sch{\"o}lkopf, B. 2013.
\newblock From ordinary differential equations to structural causal models: the deterministic case.
\newblock \emph{arXiv preprint arXiv:1304.7920}.

\bibitem[{Nauta, Bucur, and Seifert(2019)}]{nauta:19:tcdf}
Nauta, M.; Bucur, D.; and Seifert, C. 2019.
\newblock Causal discovery with attention-based convolutional neural networks.
\newblock \emph{Machine Learning and Knowledge Extraction}, 1(1).

\bibitem[{Pamfil et~al.(2020)Pamfil, Sriwattanaworachai, Desai, Pilgerstorfer, Georgatzis, Beaumont, and Aragam}]{pamfil:20:dynotears}
Pamfil, R.; Sriwattanaworachai, N.; Desai, S.; Pilgerstorfer, P.; Georgatzis, K.; Beaumont, P.; and Aragam, B. 2020.
\newblock Dynotears: Structure learning from time-series data.
\newblock In \emph{International Conference on Artificial Intelligence and Statistics}. Pmlr.

\bibitem[{Pearl(2009)}]{pearl:09:causality}
Pearl, J. 2009.
\newblock \emph{Causality}.
\newblock Cambridge university press.

\bibitem[{Peters, Janzing, and Sch{\"o}lkopf(2013)}]{peters:13:timino}
Peters, J.; Janzing, D.; and Sch{\"o}lkopf, B. 2013.
\newblock Causal inference on time series using restricted structural equation models.
\newblock \emph{Advances in neural information processing systems}, 26.

\bibitem[{Rahimi and Recht(2007)}]{rahimi:07:rfkernel}
Rahimi, A.; and Recht, B. 2007.
\newblock Random features for large-scale kernel machines.
\newblock \emph{Advances in neural information processing systems}, 20.

\bibitem[{Ramsey et~al.(2010)Ramsey, Hanson, Hanson, Halchenko, Poldrack, and Glymour}]{ramsey:10:genBIC}
Ramsey, J.~D.; Hanson, S.~J.; Hanson, C.; Halchenko, Y.~O.; Poldrack, R.~A.; and Glymour, C. 2010.
\newblock Six problems for causal inference from fMRI.
\newblock \emph{neuroimage}, 49(2).

\bibitem[{Rasmussen and Williams(2005)}]{10.5555/1162254}
Rasmussen, C.~E.; and Williams, C. K.~I. 2005.
\newblock \emph{Gaussian Processes for Machine Learning (Adaptive Computation and Machine Learning)}.
\newblock The MIT Press.

\bibitem[{Ridderbusch, Ober{-}Bl{\"{o}}baum, and Goulart(2023)}]{ridderbusch2023}
Ridderbusch, S.; Ober{-}Bl{\"{o}}baum, S.; and Goulart, P. 2023.
\newblock The past does matter: correlation of subsequent states in trajectory predictions of Gaussian Process models.
\newblock In \emph{Uncertainty in Artificial Intelligence, {UAI} 2023, July 31 - 4 August 2023, Pittsburgh, PA, {USA}}, volume 216, 1752--1761. {PMLR}.

\bibitem[{Rissanen(1983)}]{rissanen:83:integers}
Rissanen, J. 1983.
\newblock A Universal Prior for Integers and Estimation by Minimum Description Length.
\newblock \emph{AnnalsStatistics}, 11(2): 416--431.

\bibitem[{Rubenstein et~al.(2016)Rubenstein, Bongers, Sch{\"o}lkopf, and Mooij}]{rubenstein:16:dscm}
Rubenstein, P.~K.; Bongers, S.; Sch{\"o}lkopf, B.; and Mooij, J.~M. 2016.
\newblock From deterministic ODEs to dynamic structural causal models.
\newblock \emph{arXiv preprint arXiv:1608.08028}.

\bibitem[{Runge(2020)}]{runge:20:pcmci}
Runge, J. 2020.
\newblock Discovering contemporaneous and lagged causal relations in autocorrelated nonlinear time series datasets.
\newblock In \emph{Conference on uncertainty in artificial intelligence}. Pmlr.

\bibitem[{Rössler(1976)}]{rossler76}
Rössler, O. 1976.
\newblock An equation for continuous chaos.
\newblock \emph{Physics Letters A}, 57(5): 397--398.

\bibitem[{Schmidt and Lipson(2009)}]{schmidt2009freeform}
Schmidt, M.; and Lipson, H. 2009.
\newblock Distilling Free-Form Natural Laws from Experimental Data.
\newblock \emph{Science}, 324(5923).

\bibitem[{Schwarz(1978)}]{schwarz:78:bic}
Schwarz, G. 1978.
\newblock Estimating the dimension of a model.
\newblock \emph{The annals of statistics}.

\bibitem[{Strogatz(2000)}]{strogatz:2000}
Strogatz, S.~H. 2000.
\newblock \emph{Nonlinear Dynamics and Chaos: With Applications to Physics, Biology, Chemistry and Engineering}.
\newblock Westview Press.

\bibitem[{Sun, Taylor, and Bollt(2015)}]{sun:15:ocse}
Sun, J.; Taylor, D.; and Bollt, E.~M. 2015.
\newblock Causal network inference by optimal causation entropy.
\newblock \emph{SIAM Journal on Applied Dynamical Systems}, 14(1).

\bibitem[{Tsamardinos, Brown, and Aliferis(2006)}]{Tsamardinos2006TheMH}
Tsamardinos, I.; Brown, L.~E.; and Aliferis, C.~F. 2006.
\newblock The max-min hill-climbing Bayesian network structure learning algorithm.
\newblock \emph{Machine Learning}, 65.

\bibitem[{Van~der Vaart(2000)}]{van:00:book}
Van~der Vaart, A.~W. 2000.
\newblock \emph{Asymptotic statistics}, volume~3.
\newblock Cambridge university press.

\bibitem[{Vereshchagin and Vit{\'a}nyi(2004)}]{vitanyi:02:kstructurefunction}
Vereshchagin, N.~K.; and Vit{\'a}nyi, P.~M. 2004.
\newblock Kolmogorov's structure functions and model selection.
\newblock \emph{IEEETIT}, 50(12).

\bibitem[{Voortman, Dash, and Druzdzel(2010)}]{voortman:10:dbcm}
Voortman, M.; Dash, D.; and Druzdzel, M.~J. 2010.
\newblock Learning causal models that make correct manipulation predictions with time series data.
\newblock In \emph{Causality: Objectives and Assessment}. PMLR.

\bibitem[{Wang, Fleet, and Hertzmann(2005)}]{Wang2005}
Wang, J.~M.; Fleet, D.~J.; and Hertzmann, A. 2005.
\newblock Gaussian Process Dynamical Models.
\newblock In \emph{NeurIPS 2005}. Cambridge, MA, USA: MIT Press.

\bibitem[{Wang et~al.(2017)Wang, Solus, Yang, and Uhler}]{wang:igsp}
Wang, Y.; Solus, L.; Yang, K.~D.; and Uhler, C. 2017.
\newblock Permutation-based Causal Inference Algorithms with Interventions.
\newblock In \emph{NIPS}.

\bibitem[{Wendland(1995)}]{wendland:95:kernel}
Wendland, H. 1995.
\newblock Piecewise polynomial, positive definite and compactly supported radial functions of minimal degree.
\newblock \emph{Advances in computational Mathematics}, 4(1).

\bibitem[{Williams and Seeger(2000)}]{williams:00:nystrom-kernel}
Williams, C.; and Seeger, M. 2000.
\newblock Using the Nystr{\"o}m method to speed up kernel machines.
\newblock \emph{Advances in neural information processing systems}, 13.

\bibitem[{Xu, Mameche, and Vreeken(2025)}]{xu:25:topic}
Xu, S.; Mameche, S.; and Vreeken, J. 2025.
\newblock Information-Theoretic Causal Discovery in Topological Order.
\newblock In \emph{AISTATS 2025}.

\bibitem[{Xu et~al.(2022)Xu, Mian, Marx, and Vreeken}]{xu:22:heci}
Xu, S.; Mian, O.~A.; Marx, A.; and Vreeken, J. 2022.
\newblock Inferring cause and effect in the presence of heteroscedastic noise.
\newblock In \emph{ICML}. PMLR.

\end{thebibliography}
\clearpage
\makeatletter
\@ifundefined{isChecklistMainFile}{
  \newif\ifreproStandalone
  \reproStandalonetrue
}{
  \newif\ifreproStandalone
  \reproStandalonefalse
}
\makeatother

\ifreproStandalone
\documentclass[letterpaper]{article}
\usepackage[submission]{aaai2026}
\setlength{\pdfpagewidth}{8.5in}
\setlength{\pdfpageheight}{11in}
\usepackage{times}
\usepackage{helvet}
\usepackage{courier}
\usepackage{xcolor}
\frenchspacing

\begin{document}
\fi
\setlength{\leftmargini}{20pt}
\makeatletter\def\@listi{\leftmargin\leftmargini \topsep .5em \parsep .5em \itemsep .5em}
\def\@listii{\leftmargin\leftmarginii \labelwidth\leftmarginii \advance\labelwidth-\labelsep \topsep .4em \parsep .4em \itemsep .4em}
\def\@listiii{\leftmargin\leftmarginiii \labelwidth\leftmarginiii \advance\labelwidth-\labelsep \topsep .4em \parsep .4em \itemsep .4em}\makeatother

\setcounter{secnumdepth}{0}
\renewcommand\thesubsection{\arabic{subsection}}
\renewcommand\labelenumi{\thesubsection.\arabic{enumi}}

\newcounter{checksubsection}
\newcounter{checkitem}[checksubsection]

\newcommand{\checksubsection}[1]{%
  \refstepcounter{checksubsection}%
  \paragraph{\arabic{checksubsection}. #1}%
  \setcounter{checkitem}{0}%
}

\newcommand{\checkitem}{%
  \refstepcounter{checkitem}%
  \item[\arabic{checksubsection}.\arabic{checkitem}.]%
}
\newcommand{\question}[2]{\normalcolor\checkitem #1 #2 \color{blue}}
\newcommand{\ifyespoints}[1]{\makebox[0pt][l]{\hspace{-15pt}\normalcolor #1}}

\section*{Reproducibility Checklist}

\vspace{1em}
\hrule
\vspace{1em}

\textbf{Instructions for Authors:}

This document outlines key aspects for assessing reproducibility. Please provide your input by editing this \texttt{.tex} file directly.

For each question (that applies), replace the ``Type your response here'' text with your answer.

\vspace{1em}
\noindent
\textbf{Example:} If a question appears as
\begin{center}
\noindent
\begin{minipage}{.9\linewidth}
\ttfamily\raggedright
\string\question \{Proofs of all novel claims are included\} \{(yes/partial/no)\} \\
Type your response here
\end{minipage}
\end{center}
you would change it to:
\begin{center}
\noindent
\begin{minipage}{.9\linewidth}
\ttfamily\raggedright
\string\question \{Proofs of all novel claims are included\} \{(yes/partial/no)\} \\
yes
\end{minipage}
\end{center}
Please make sure to:
\begin{itemize}\setlength{\itemsep}{.1em}
\item Replace ONLY the ``Type your response here'' text and nothing else.
\item Use one of the options listed for that question (e.g., \textbf{yes}, \textbf{no}, \textbf{partial}, or \textbf{NA}).
\item \textbf{Not} modify any other part of the \texttt{\string\question} command or any other lines in this document.\\
\end{itemize}

You can \texttt{\string\input} this .tex file right before \texttt{\string\end\{document\}} of your main file or compile it as a stand-alone document. Check the instructions on your conference's website to see if you will be asked to provide this checklist with your paper or separately.

\vspace{1em}
\hrule
\vspace{1em}


\checksubsection{General Paper Structure}
\begin{itemize}

\question{Includes a conceptual outline and/or pseudocode description of AI methods introduced}{(yes/partial/no/NA)}
yes

\question{Clearly delineates statements that are opinions, hypothesis, and speculation from objective facts and results}{(yes/no)}
yes

\question{Provides well-marked pedagogical references for less-familiar readers to gain background necessary to replicate the paper}{(yes/no)}
yes

\end{itemize}
\checksubsection{Theoretical Contributions}
\begin{itemize}

\question{Does this paper make theoretical contributions?}{(yes/no)}
yes

	\ifyespoints{\vspace{1.2em}If yes, please address the following points:}
        \begin{itemize}
	
	\question{All assumptions and restrictions are stated clearly and formally}{(yes/partial/no)}
	yes

	\question{All novel claims are stated formally (e.g., in theorem statements)}{(yes/partial/no)}
	yes

	\question{Proofs of all novel claims are included}{(yes/partial/no)}
	yes

	\question{Proof sketches or intuitions are given for complex and/or novel results}{(yes/partial/no)}
	yes

	\question{Appropriate citations to theoretical tools used are given}{(yes/partial/no)}
	yes

	\question{All theoretical claims are demonstrated empirically to hold}{(yes/partial/no/NA)}
	partial

	\question{All experimental code used to eliminate or disprove claims is included}{(yes/no/NA)}
	yes
	
	\end{itemize}
\end{itemize}

\checksubsection{Dataset Usage}
\begin{itemize}

\question{Does this paper rely on one or more datasets?}{(yes/no)}
yes

\ifyespoints{If yes, please address the following points:}
\begin{itemize}

	\question{A motivation is given for why the experiments are conducted on the selected datasets}{(yes/partial/no/NA)}
	yes

	\question{All novel datasets introduced in this paper are included in a data appendix}{(yes/partial/no/NA)}
	yes

	\question{All novel datasets introduced in this paper will be made publicly available upon publication of the paper with a license that allows free usage for research purposes}{(yes/partial/no/NA)}
	yes

	\question{All datasets drawn from the existing literature (potentially including authors' own previously published work) are accompanied by appropriate citations}{(yes/no/NA)}
	NA

	\question{All datasets drawn from the existing literature (potentially including authors' own previously published work) are publicly available}{(yes/partial/no/NA)}
	NA

	\question{All datasets that are not publicly available are described in detail, with explanation why publicly available alternatives are not scientifically satisficing}{(yes/partial/no/NA)}
	NA

\end{itemize}
\end{itemize}

\checksubsection{Computational Experiments}
\begin{itemize}

\question{Does this paper include computational experiments?}{(yes/no)}
yes

\ifyespoints{If yes, please address the following points:}
\begin{itemize}

	\question{This paper states the number and range of values tried per (hyper-) parameter during development of the paper, along with the criterion used for selecting the final parameter setting}{(yes/partial/no/NA)}
	yes

	\question{Any code required for pre-processing data is included in the appendix}{(yes/partial/no)}
	yes

	\question{All source code required for conducting and analyzing the experiments is included in a code appendix}{(yes/partial/no)}
	yes

	\question{All source code required for conducting and analyzing the experiments will be made publicly available upon publication of the paper with a license that allows free usage for research purposes}{(yes/partial/no)}
	yes
        
	\question{All source code implementing new methods have comments detailing the implementation, with references to the paper where each step comes from}{(yes/partial/no)}
	partial

	\question{If an algorithm depends on randomness, then the method used for setting seeds is described in a way sufficient to allow replication of results}{(yes/partial/no/NA)}
	yes

	\question{This paper specifies the computing infrastructure used for running experiments (hardware and software), including GPU/CPU models; amount of memory; operating system; names and versions of relevant software libraries and frameworks}{(yes/partial/no)}
	yes

	\question{This paper formally describes evaluation metrics used and explains the motivation for choosing these metrics}{(yes/partial/no)}
	yes

	\question{This paper states the number of algorithm runs used to compute each reported result}{(yes/no)}
	yes

	\question{Analysis of experiments goes beyond single-dimensional summaries of performance (e.g., average; median) to include measures of variation, confidence, or other distributional information}{(yes/no)}
	yes

	\question{The significance of any improvement or decrease in performance is judged using appropriate statistical tests (e.g., Wilcoxon signed-rank)}{(yes/partial/no)}
	yes

	\question{This paper lists all final (hyper-)parameters used for each model/algorithm in the paper’s experiments}{(yes/partial/no/NA)}
	partial

\end{itemize}
\end{itemize}
\ifreproStandalone
\end{document}
\fi

\clearpage
\appendix
\onecolumn
\setcounter{equation}{0} 
\renewcommand{\theequation}{A.\arabic{equation}} 
\setcounter{secnumdepth}{1}
\section{Theoretical Results}\label{appendix:theoretical_results}

\setcounter{theorem}{0}
\setcounter{assumption}{0}
\setcounter{definition}{0}
\renewcommand{\thedefinition}{\thesection.\arabic{definition}}

In this section we provide proofs for our theoretical results. For ease of recall, we restate Assumptions. 1,2 and Theorem. 1 below along with the definitions of regularized log-likelihood scores and specifically the BIC score~\cite{schwarz:78:bic}.

\begin{definition}[Bayesian Information Criterion]\label{def:bic}
For a statistical model $\mathcal{M}_k$ with $k$ free parameters and maximum log-likelihood $\ell(\hat{\theta}_k)$, the \textit{Bayesian Information Criterion (BIC)} is defined as
\begin{equation}
\mathrm{BIC}(\mathcal{M}_k) = -2\,\ell(\hat{\theta}_k) + k \log N,
\end{equation}
where $N$ is the number of observations. Lower values of $\mathrm{BIC}$ indicate a preferred model under an asymptotic approximation to the Bayes factor \citep{schwarz:78:bic}.
For a Gaussian likelihood with residual sum of squares $\mathrm{RSS}$, this further simplifies to
\begin{equation}\label{eq:gBIC}
\mathrm{BIC} = N \log\!\left(\frac{\mathrm{RSS}}{N}\right) + k \log N \; ,
\end{equation}
{with generalized BIC taking the form
\begin{equation}\label{eq:gBIC_2}
    \mathrm{BIC} = a_1\cdot N \log\!\left(\frac{\mathrm{RSS}}{N}\right) + a_2\cdot k \log N \;,
\end{equation}
with $a_1$ and $a_2$ as constant terms independent of $N$~\cite[Appendix. A]{ramsey:10:genBIC}.}
\end{definition}
\noindent
In the following we focus on an arbitrary submodel. We partially skip indices identifying the specific submodel where unnecessary. 

\begin{definition}[Regularized log-likelihood score] \label{de:rlls}
    Given $N$ observations, an empirical estimate of residual noise variance, $\hat{\sigma}^2$, and model dependent constants $c_0,c_1,c_2$ independent of $N$, we define a regularized log-likelihood score 
    \begin{equation*}
        RLLS(N,\hat{\sigma}^2) = c_0 \cdot N \cdot log(\hat{\sigma}^2) + c_1\kappa(N) + c_2
    \end{equation*}
    with $\kappa$ a function that fullfills $\kappa(N) < N$. 
\end{definition}

\begin{assumption}[Finite dimensions]
$K$ is finite-dimensional. 
\end{assumption}

\noindent {
\begin{assumption}[Bounded hyperparameters and precision]
The lengtscale parameter $\alpha_i$, the variance parameter $\beta_i$ are upper bounded. All precisions $|\rho_i|$ within each model $M_j$ are upper bounded.
\end{assumption}}

\begin{lemma}
   Given a DSCM $\mathcal{S}$, let $\trajectory$ be a trajectory generated from $\mathcal{S}$ and let $\bar{L}(\trajectory,M_j) = L(M_j) + L(\nu_{j}) - C$. If Assms.~\ref{assm:finite-support},\ref{assm:boundedprecision} hold, it holds asymptotically
    \begin{equation*}
     \bar{L}(\trajectory,M_j) \leq c_0 \cdot N \cdot log(\hat{\sigma}_{\color{purple}{j}}^2) +   c_1^{(j)}  log(N) + c_2^{(j)}.
    \end{equation*}
    with constants $c_0,c_1^{(j)},c_2^{(j)}$ independent of $N$. 
\end{lemma}
\begin{proof}
\noindent
Recall that we formally define $C$ as:
\begin{equation*}
C = \log \left(\frac{1}{r_\lambda}\right) \cdot \frac{N(N-1)}{2} +  2||\theta||  + \frac{N}{2}\left( \frac{1}{\ln2} + \log(2\pi)\right)\;.
\end{equation*}

~\\~\\
\noindent
Specifically, whenever we subtract $C$ from subsequent expressions, we subtract the first term from Eq.~\eqref{eq:LF}, second term from Eq.~\eqref{eq:encode-params} and the last term from Eq.~\eqref{eq:residuals} contained in those expressions. We equivalently obtain,
\begin{equation*}
    \bar{L}(\trajectory,M_j)=   L(M_j) + L(\nu_j)  - C = \bar{L}(M_j) +  \bar{L}(\nu_j) \;,
\end{equation*}
with $\bar{L}(M_j)$ resp. $\bar{L}(\nu_j)$ defined formally as,
\begin{equation*}
     \bar{L}(M_j) =  {L}(M_j) - C_1, 
\end{equation*}
\begin{equation*}
    \bar{L}(\nu_j) = {L}(\nu_j) - C_2 \;,
\end{equation*}
where $C_1 = \log \left(\frac{1}{r_\lambda}\right) \cdot \frac{N(N-1)}{2} +  2||\theta||$ and $C_2 = \frac{N}{2}\left( \frac{1}{\ln2} + \log(2\pi)\right)$.

\noindent
We begin by looking at the remaining term in Eq.~\eqref{eq:residuals} we observe that it can be rewritten as
\begin{equation}\label{eq:datacost}
    \bar{L}(\nu_j) = \frac{N}{2}   log( \hat{\sigma_j}^2 )    
    =    c_0 \cdot N \cdot \log(\hat{\sigma_j}^2)   \; ,
\end{equation}
where the constant $c_0 = \frac{1}{2}$ is independent of the {{sub}}model.

\noindent
Next, we analyze the local model cost $\bar{L}(M_j)$ and show that it does not grow faster than the likelihood term in Eq.\eqref{eq:residuals}. \\

\noindent
For local models it holds that
\begin{equation}
\label{eq:constcost}
 L_{\mathbb{N}}(||\pa_{{j}}||) + ||\pa_{{j}}||\log D = c_3^{({{j}})}
\end{equation}
as these two terms are independent of $N$. This leaves us with Eq.~\eqref{eq:encode-params} without the $||\theta||$ term that we excluded from $L$ to get $\bar{L}$ {{where we omit the submodel index $j$ for ease of notation}}:
\begin{equation}
\bar{L}_{p}(\theta) = \sum_{i = 1}^{||\theta||} L_{\mathbb{N}}(\lceil \theta_i \cdot 10^{\rho_i} \rceil) + L_{\mathbb{N}}(|\rho_i|)\; 
\end{equation}
Referring to eigenvalues as $\lambda$ with size $||\lambda||$ and using the notation $\gamma_i = (\alpha_i, \beta_i)$ for length scales and variances, $||\gamma||$ for their number, let $\theta$ with size $||\theta|| = ||\gamma|| + ||\lambda||$ be the set of values assigned to each eigenvalue and parameter. Letting $\rho^* = \max (|\rho_i|)$, we have,

\begin{align}
    \bar{L}_{p}(\theta) &= \bar{L}_{p}(\lambda) + \bar{L}_{p}(\gamma)\\
     &= \bar{L}_{p}(\lambda) + c_4\\
     &= \sum_{i = 1}^{||\lambda||} \left(L_{\mathbb{N}}(\lceil \lambda_i \cdot 10^{\rho_i} \rceil) + L_{\mathbb{N}}(|\rho_i|) \right)  + c_4\\
     &\leq \sum_{i = 1}^{||\lambda||} \left(L_{\mathbb{N}}(\lceil \lambda_i \cdot 10^{\rho^*} \rceil) + L_{\mathbb{N}}(\rho^*) \right)  + c_4\label{eq:lceil1}\\
     &< \sum_{i = 1}^{||\lambda||} \left(2 \log(\lambda_i \cdot 10^{\rho^*}) + 2 \log(\rho^*) \right)  + c_4\label{eq:l2log1}\\
     &= \sum_{i = 1}^{||\lambda||} 2 \log(\rho^*\cdot \lambda_i \cdot 10^{\rho^*}) + c_4\\
    &= \sum_{i = 1}^{||\lambda||} 2 \log(c_5 \lambda_i) + c_4\\
    &< \sum_{i = 1}^{||\lambda||} 2 \log(1 + c_5 \lambda_i) + c_4\label{eq:logsum}\\
    &= 2 \cdot \log |\mathbf{I}+ c_5\mathbf{K}|~\label{eq:kakade} +  c_4
\end{align}
where we transition from Eq.~\eqref{eq:lceil1} to Eq.~\eqref{eq:l2log1} by using the inequality $L_\mathbb{N}(\lceil x \rceil) < 2 \log x$ for all $x\geq10$, which is always the case for us as we would use $L_p$ with atleast $p=1$. Eq.~\eqref{eq:kakade} implies that the complexity of $L_{p}$ is upper bounded by the complexity of the log determinant term of the learned Kernel matrix $\mathbf{K}$. To evaluate this complexity, we include the result of~\citet[Sec 3.2]{kakade:05:gp-bounds} which states that for Kernels with finite dimensions (ref. Assm 1) we have,
\begin{equation}
    \log |\mathbf{I}+ {c_5}\mathbf{K}| \leq D \log{\left(1+\frac{c_{5} N}{D} \right) }
\end{equation}
Substituting this into Eq.~\eqref{eq:kakade} {{and re-introducing the submodel index $j$}}, we get
\begin{equation}\label{eq:modcost}
    \bar{L}_{p}(\theta) < c_1^{(j)}\cdot \log N +   c_6^{(j)} \;.
\end{equation}
This means
\begin{equation*}
    \bar{L}_{p}(\theta) = \mathcal{O}(\log N).
\end{equation*}  

\noindent
Combining Eq.~\eqref{eq:datacost}, Eq.~\eqref{eq:constcost} and Eq.~\eqref{eq:modcost} we get
\begin{equation}
     \bar{L}(\trajectory,M_j) \leq c_0 \cdot N \cdot log(\hat{\sigma_j}^2) +   c_1^{(j)}  log(N) + c_2^{(j)}.
\end{equation}

\noindent
Note that the second term is in  $\mathcal{O}(\log N)$ which means that the upper bound of the parameter penalty grows sublinearly in $N$ whereas the likelihood term scales in $N$ and dominates in the limit, giving us a valid regularized log-likelihood score.
\end{proof}

\begin{theorem}
   Given a DSCM $\mathcal{S}$, let $\trajectory$ be a trajectory generated from $\mathcal{S}$ and let $\bar{L}(\trajectory,M) =  \sum_{j=1}^{D} \bar{L}(\trajectory,M_j)$. If Assms.~\ref{assm:finite-support},\ref{assm:boundedprecision} hold, it holds asymptotically.
    \begin{equation*}
     \bar{L}(\trajectory,M) \leq c_0 \cdot N \cdot log(\hat{\sigma}^2) +   c_1  log(N) + c_2.
    \end{equation*}
    with constants $c_0,c_1$ and $c_2$ independent of $N$ and $\bar{L}(\trajectory,M)$ asymptotically is a valid regularized log-likelihood score according to Definition \ref{de:rlls}. 
\end{theorem}
\begin{proof}
We first split $\bar{L}(\trajectory,M)$ into two sums as formalized in the proof of Lemm.~\ref{lemm:mec}, 
\begin{equation*}
    \bar{L}(\trajectory,M)= \sum_{j=1}^D  \bar{L}(\nu_j)  + \sum_{j=1}^D \bar{L}(M_j) \;.
\end{equation*}

Next we consider these two terms individually. Note that we can expand the first term as
\begin{align*}
    \sum_{j=1}^D \bar{L}(\nu_j) &= \sum_{j=1}^D c_0 \cdot N \cdot \log(\hat{\sigma_j}^2)  \\
    &= c_0 \cdot N \cdot\sum_{j=1}^D  \log(\hat{\sigma_j}^2)  \\
    &= c_0 \cdot N \cdot \log(\prod_{j=1}^D\hat{\sigma_j}^2)  \\
    &= c_0 \cdot N \cdot \log(\hat{\sigma}^2)  \\
\end{align*}

We expand the second term similarly,
\begin{align*}
    \sum_{j=1}^D \bar{L}(M_j) &\leq \sum_{j=1}^D \left( c_1^{(j)}\cdot \log N +   c_2^{(j)}  \right) \\
    &= \log N \cdot \sum_{j=1}^D c_1^{(j)} + \sum_{i=1}^D c_2^{(j)} \\
    &= c_1\cdot \log N + c_2 \;,\\
\end{align*}
with where we combine the sum $\sum_{j=1}^D c_1^{(j)}$ into the constant $c_1$ and $c_2 = \sum_{i=1}^D c_2^{(j)}$.

\noindent
Putting together the two results we get,

 \begin{equation*}
     \bar{L}(\trajectory,M) \leq c_0 \cdot N \cdot log(\hat{\sigma}^2) +   c_1  log(N) + c_2.
\end{equation*}

\end{proof}

\begin{corollary}
    Model selection using ${L}(\trajectory,M)$ is equivalent to model selection using $\bar{L}(\trajectory,M)$.
\end{corollary}
\begin{proof} 
Note that even though $L(\trajectory,M)$ contains terms in $\mathcal{O}(N^2)$ i.e. $\log \left(\frac{1}{r_\lambda}\right) \cdot \frac{N(N-1)}{2} \;,$ (Eq.\eqref{eq:LF}) and in Eq.~\eqref{eq:encode-params} we have $2||\theta|| = 2N$, these terms remain constant across models fitted on same length $\trajectory$. 
The same holds for the terms $\frac{N}{2}\left( \frac{1}{ln2} + log(2\pi)\right)$ and $L_{global}(M)$.Therefore, they do not influence the model being learned. Hence, the $N^2$ term of Eq.\eqref{eq:LF}, $2||\theta||$ and $\frac{N}{2}\left( \frac{1}{ln2} + log(2\pi)\right)$ and $L_{global}(M)$ cancel out when comparing two models using $L(\trajectory,M)$ and do not affect the score change and do not change the model selection process.
\end{proof}

~\\
Regularized log-likelihood scores have been widely used to establish consistency~\cite{huang:18:gges,marx:19:sloppy,brouillard:20:dcdi,mian:21:globe,xu:22:heci,xu:25:topic}. Both the Bayesian Information Criterion (BIC) and its generalization introduced in Def.~\ref{def:bic} can be viewed as special cases of regularized log-likelihood scores. These criteria are well established for independent and identically distributed (i.i.d.) data; see, e.g., \cite[Sec.~5.5]{van:00:book}, \cite{kashyap:82:optimal,haughton:88:curveexpBIC,kass:95:BICAnalysis,chickering:02:ges,marx:19:sloppy,lu:21:optimalsearch,mian:21:globe}. For BIC-type scores, consistency guarantees have been shown to extend to non-i.i.d. time-series data under the following conditions:
\begin{enumerate}[label=\roman*.,leftmargin=4.5em,rightmargin=5.2em,itemsep=0pt]
    \item \label{cavanaugh} The likelihood term satisfies mild regularity conditions~\cite[Sec.~3]{cavanaugh:99:sic}.
    \item \label{claesken} The true model is contained in the candidate model set, and the number of models does not grow with $N$~\cite[Thm~4.2]{claeskens:08:book}.
\end{enumerate}
Condition~\ref{cavanaugh} involves standard assumptions such as twice continuous differentiability of the likelihood function with respect to its parameters, which typically holds for Gaussian process regression. Condition~\ref{claesken} follows directly from causal sufficiency and Assumption~1 in our setting. \\[12pt]
The form of regularized log-likelihood scores we describe is general, and our particular instantiation represents one specific case within this broader class. Most of the cited works analyze alternative forms of regularized log-likelihoods and establish convergence results. 
Because our score includes a term of the form $N\log(\sigma^2)$ plus a correction term, we are optimistic that convergence guarantees or selection of the correct model up to precision $\rho^*$ can be formally shown in future work. \\[12pt]

\clearpage
\twocolumn

\section{Benchmarks Details}\label{appendix:benchmarks}
Hereby, we provide further details on our experimental setting.

\subsection{Computational Resources} All our results are efficiently run by parallelizing the greedy search on a AMD EPYC 7643 CPU with 60GB RAM allocated.

\subsection{Physical Models} \label{appendix:physical_models}

We provide a detailed description for each benchmark, including the underlying physical models and their initial conditions. Each dynamics described below has been integrated using $RK4(5)$, after which an additive gaussian noise $\epsilon \sim \mathcal{N}(0, \sigma)$ has been applied. 

\paragraph{Empty Causal Graph:} For the empty causal graph, we generate $D = 4$ independent trajectories using $D$ ODEs that do not have any inter-dependency. Those ODEs are in the form $\dfrac{dy}{dt} = e^{at}$ with $a \sim \mathcal{U}([-2.5, 2.5])$. Those ODEs result in either an exponential growth or decay. The timeline of integration is $T = [0, 40]$ with $\Delta t = 0.1$, what has been tested with regular and irregular $(20\%)$ timelines. The additive nosie std. is $\sigma = 0.05$.

\paragraph{Diamond Causal Graph:}We test on the Diamond Causal Graph, which represents a simple benchmark containing all the basic causal structures: The fork, the mediator, and the collider. Below, we provide an illustration.

\begin{equation}\label{fig:diamond_graph}
\vcenter{\hbox{\includegraphics[width=0.28\linewidth]{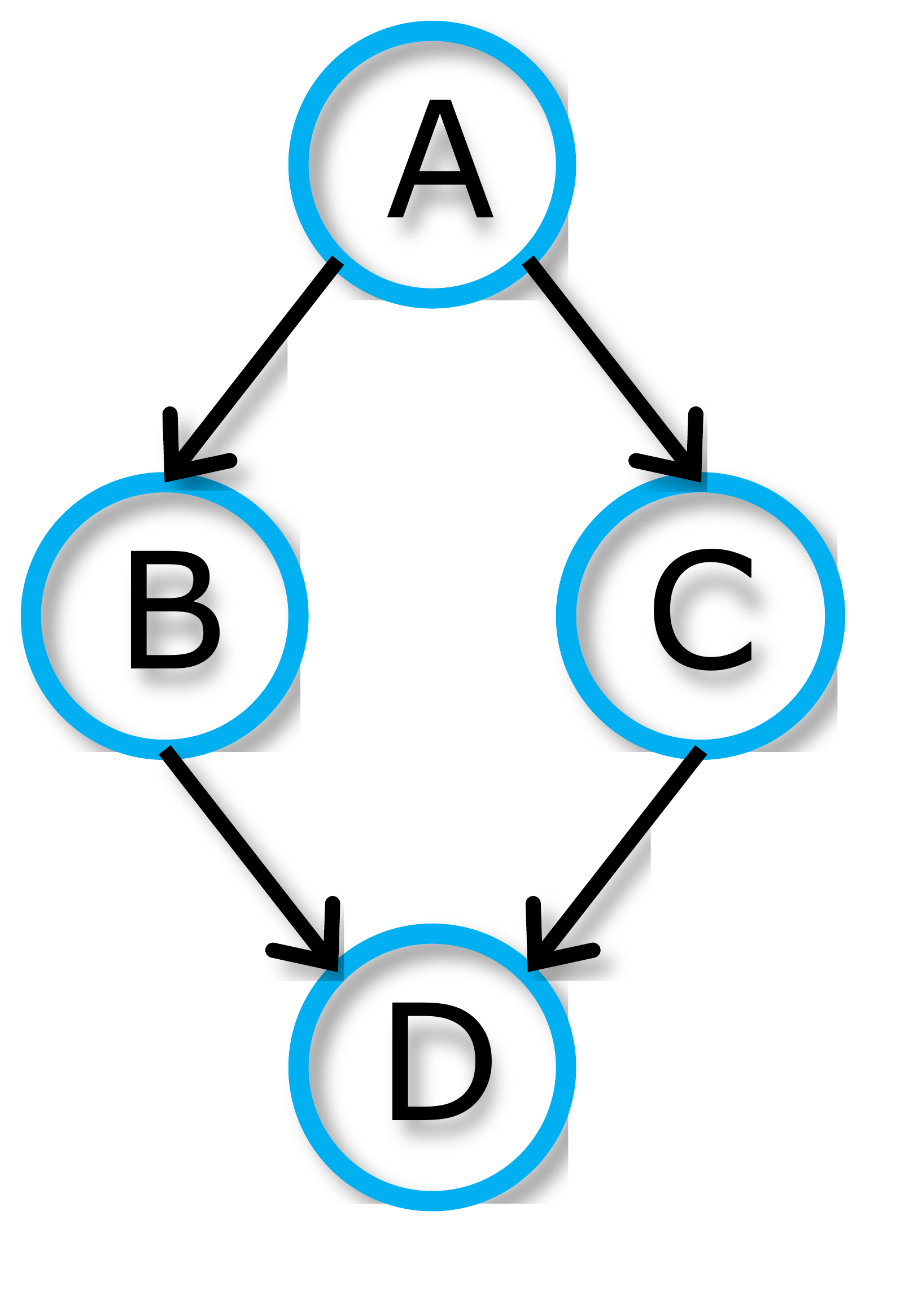}}}
\qquad\qquad
\begin{aligned}
\frac{dA}{dt} &= \dfrac{1}{2} \cdot \sin{(t)} \\
\frac{dB}{dt} &= \dfrac{1}{2} \cdot \sin{(A)} \\
\frac{dC}{dt} &= - 2 \cdot \cos{(A)}\\
\frac{dD}{dt} &= B + C\\
\end{aligned}
\end{equation}
The initial conditions are randomly sampled from a gaussian distribution with $\mu = 0$ and $\sigma = 1$. The timeline of integration is $T = [0, 10]$ and $\Delta t = 0.05$. We tested on both a regular and irregular ($20\%$) timeline. For the Diamond, the additive noise std. is $\sigma  = 0.005$.

\paragraph{Erdős–Rényi Random DAG:} We simulate Erdős–Rényi random DAGs with graph size $D=\{ 5, 10, 15\}$ and probability of a directed edge appearing (density parameter) as $p = 2 / D$. We sample a set of ODEs following the dependencies induced by the causal graph as written below:
\begin{equation}\label{eq:erdos_renyi_equations}
\frac{dX_i}{dt} = \sin{\Bigg(\sum_{k = 0}^{|Pa_i|} \alpha_{ik} \cdot [Pa_i]_k \Bigg)}  \quad \text{with }  \alpha_{ik}\sim \mathcal{U}([1, 2])
\end{equation}
Where $[Pa_i]_k$ is the k-th parent of $X_i$. For root-nodes, instead, we insert a time-dependent ODE $\frac{dX_i}{dt} = \sin{(\alpha_i t)}$. Initial conditions are randomly sampled from a gaussian distribution with $\mu = 0$ and $\sigma = 1$. The additive noise has standard deviation $\sigma = 0.005$.

\paragraph{Cyclic Erdős–Rényi Random Graphs:} When considering the cyclic version of Erdős–Rényi graphs, we first generate a random DAG using the procedure described in the paragraph above, to which we systematically add 2 cycles. Each cycle is added by randomly selecting a node (one which is not a root-node), and adding a directed connection to one of its ancestors (randomly chosen). The ODE system for this dynamical system follows the same logic described above.

\paragraph{Double-Mass Spring System}
We simulate the following ODE modeling a double-mass spring system. 
\begin{align}
m_1 \frac{d^2y_1}{d^2t}&= f_1(t) + k_2 (y_2 - y_1) - k_1 y_1\\
m_2 \frac{d^2y_2}{d^2t}&= f_2(t) - k_2 (y_2 - y_1)
\end{align}
Where $k_1$ and $k_2$ are the elastic constants of each spring, and $m_1, m_2$ are the masses. We simulate it by setting $k_1 = k_2 = 1$, and set masses equal to $m_1 = m_2 = 1$. Further, we simulate an autonomous ODE, therefore we eliminate any damping factor by setting $f(t) = 0$. The system can be rewritten as a system of linear ODEs:
\begin{align}
\frac{dv_1}{dt} &= y_1 \\
\frac{dv_2}{dt} &= y_2 \\
\frac{dy_1}{dt} &= - 2 v_1 + v_2\\
\frac{dy_2}{dt} &= v_1 - v_2
\end{align}
We sample initial conditions from a gaussian distribution with $\mu = 0$ and $\sigma = 1$. The additive noise has standard deviation $\sigma = 0.005$. The timeline of integration is $T = [0, 15]$ and $\Delta t = 0.1$.

\paragraph{Rössler Oscillator} The Rössler Oscillator \citep{rossler76} is a hyper-chaotic system. We perform our experiments with $D = 10$, and the set of ODEs is the following
\begin{align}
\frac{dx_1}{dt} &= ax_1(t) - x_2(t) \\
\frac{dx_i}{dt} &= sin(x_{i - 1}(t)) - sin(x_{i +2}(t)) \\
\frac{dx_D}{dt} &= \epsilon + b x_D (t) \cdot (x_{D-1} (t) - q)
\end{align}
Initial conditions are sampled from a gaussian distribution with $\mu = 0$ and $\sigma = 1$. The additive noise has standard deviation $\sigma = 0.001$. The timeline of integration is $T = [0, 40]$ and $\Delta t = 0.1$. The parameters are chosen to make the system behave hyper-chaotically, i.e., $a = 0$, $\epsilon = 0.1$, $b=4$ and $q = 2$.

\paragraph{Real Double-Linear System} In \citep{schmidt2009freeform}, authors provide accurate trajectories of a real double-mass spring system via motion-tracking. The dynamics is the same as the Double-Mass Spring system, and the ground truth graph is also the same.

\subsection{Generation of Timelines} \label{appendix:timelines}

Although our goal is to learn a continuous-time system which is defined over a dense interval $T$, in practice we only have access to a finite amount of samples at specific time instants. Those timelines are regular when trajectory points are sampled over constant time interval. In detail, we define regular timelines as those timelines with constant stepsize, i.e.\, $\Delta t = t_i - t_{i-1} $ is constant $\forall i = 2, \dots, N$.

When $\Delta t$ changes over time, we say that the timeline is irregular. Our timelines are generated by having an average $\Delta t$ which can vary on an interval which can be as wide as a specific percentage of its value. Mathematically,
\[
t_{i+1} = t_i + \Delta t \Big(1 + (w - \dfrac{1}{2})b\Big) \quad \text{with } w \sim \mathcal{U}(0,1),
\]
where $b \in [0,1]$ controls the amount of irregularity. For example, $b = 0.5$ means that the average step-size $\Delta t$ can vary at most $50\%$ of its absolute value. For average $\Delta t = 1$ and $20\%$ irregularity, the actual step-size can vary within $0.8$ and $1.2$.

\subsection{Additional Results}\label{appendix:additional_results}

In this section, we repeat all our experiments while using \ourmethod with GPs using Polynomial kernels in addition to RBF ones. We show the statistics in Figure n:~\ref{fig:polystats}. For the sanity check on the empty graph, we show the results in Table n:~\ref{tab:empty}. We show the results for the diamond structure In Table n:~\ref{tab:diamond}. The metrics computed on 2-mass spring system are shown in Table n:~\ref{tab:appmass}
\renewcommand{\thetable}{\thesection.\arabic{table}} 
\renewcommand{\thefigure}{\thesection.\arabic{figure}} 
\setcounter{figure}{0} 
\setcounter{table}{0} 

\begin{figure}[t]
	\includegraphics[width=0.98\linewidth]{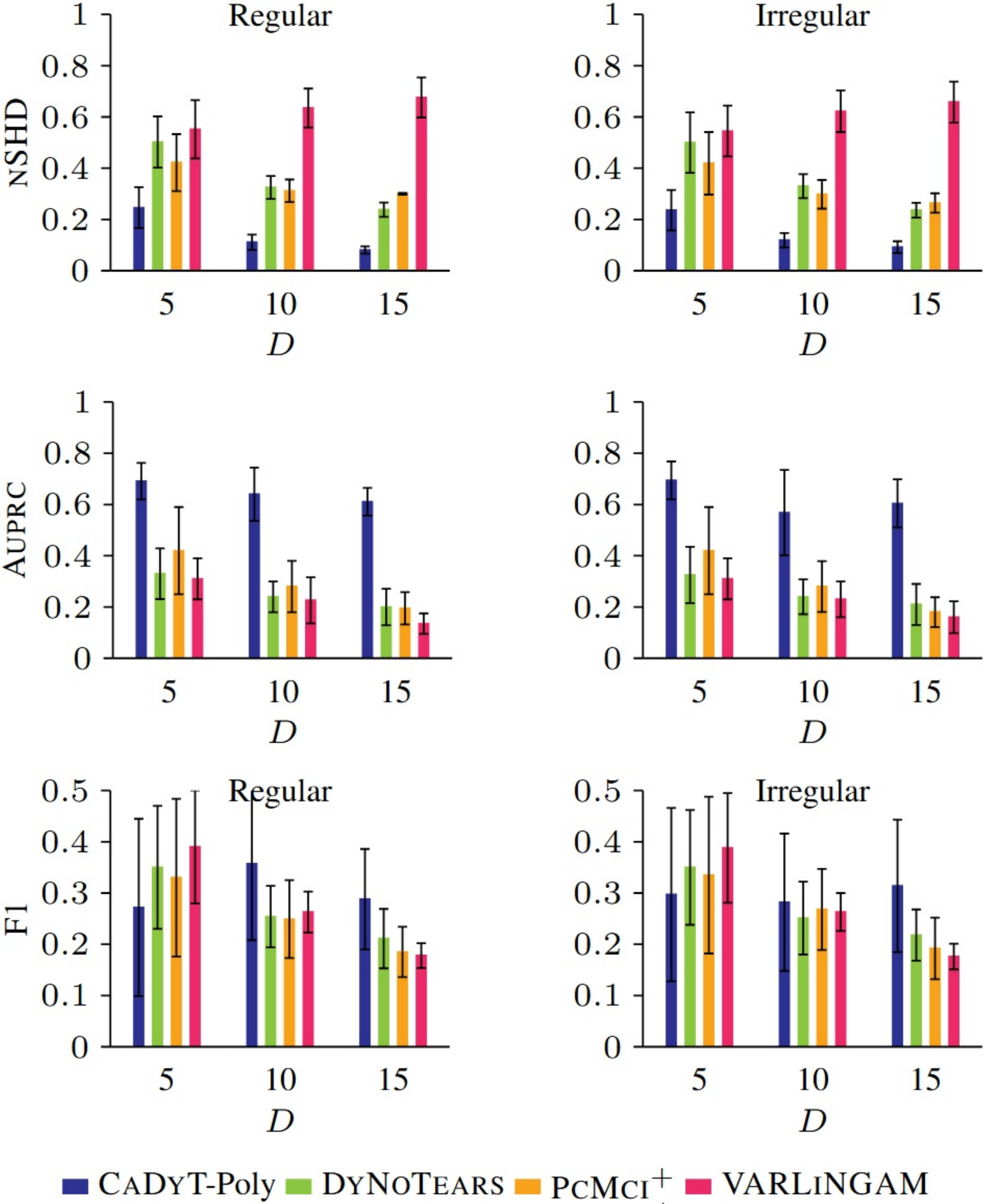}
    \caption{[\textsc{CaDyT} with Polynomial Kernel] \nSHD (top, lower is better), \textsc{Auprc} (middle, higher is better) \fone (bottom, higher is better) for random graphs of sizes $D \in \{5,10,15 \}$ for regularly sampled data (left) and irregularly sampled data (right).}
    \label{fig:polystats}
\end{figure}

\begin{table}[t!]
\centering
\caption{\textsc{\nSHD} (Lower is better) and \aupr (Higher is better) for simulations on 2-mass spring system. The ground truth contains four variables and six edges. \ourmethod ranks highest for both metrics.}
\label{tab:springmass}
\begin{tabular}{ccc}
\hline
\textbf{Method} & \textbf{\nSHD}$\downarrow$ & \textbf{AUPRC}$\uparrow$ \\ \hline
\dynotears        & 0.67            & 0.22           \\
\pcmci            & 0.62            & 0.24           \\
\varlingam        & 0.50            & 0.39           \\ 
\ourmethod (ours) & \textbf{0.25}   & \textbf{0.79}  \\ \hline
\end{tabular}
\end{table}

\begin{table*}[t]
\centering
\begin{tabular}{ccc}
\hline
\textbf{Method} & \textbf{\nSHD}$\downarrow$  \\ \hline
\dynotears        & 0.25(0.00)     \\
\pcmci            & 0.04(0.04)     \\
\varlingam        & 0.02(0.04)     \\ \hline
\ourmethod (Poly) & \textbf{0.00(0.00)}  \\ 
\ourmethod (RBF) & \textbf{0.00(0.00)}  \\ \hline
\end{tabular}
\caption{\textsc{\nSHD} (Lower is better) for simulations on the empty causal graph. The ground truth contains four variables and zero edges. Results are rounded to two decimal places. \ourmethod ranks best, being capable of sensing the absence of causal relationships with both the RBF and the Polynomial Kernel, with every integrator (AB1, AB2, AB3).}\label{tab:empty}
\end{table*}

\begin{table*}[t!]
\centering
\begin{tabular}{ccccc}
\hline 
 & \multicolumn{2}{c}{Regular} & \multicolumn{2}{c}{Irregular} \\\hline
\textbf{Method} & \textbf{\nSHD}$\downarrow$ & \textbf{AUPRC}$\uparrow$ & \textbf{\nSHD}$\downarrow$ & \textbf{AUPRC}$\uparrow$ \\ \hline
\dynotears        & 0.694(0.074) & 0.168(0.023) & 0.650(0.094)           & 0.184(0.033) \\
\pcmci            & 0.450(0.039)  & 0.132(0.023)  & 0.462(0.462)            & 0.139(0.030)  \\
\varlingam        & 0.668(0.072) & 0.174(0.009)  & 0.612(0.064)           & 0.181(0.013)        \\ 
\ourmethod (RBF) & 0.118(0.062)
  & 0.834(0.156) & 0.112(0.057) & 0.866(0.109) \\
\ourmethod (Poly) & \textbf{0.106(0.083)}  & \textbf{0.883(0.199)} & \textbf{0.087(0.073)}            & \textbf{0.925(0.191)} \\\hline
\end{tabular}
\caption{\nSHD (Lower is better) and \aupr (Higher is better) for simulations on the diamond causal graph. The ground truth contains four variables and four edges.}\label{tab:diamond}
\end{table*}

\begin{table*}[t!]
\centering
\begin{tabular}{ccccc}
\hline 
 & \multicolumn{2}{c}{Synthetic Double-Mass} & \multicolumn{2}{c}{Real Double-Linear} \\\hline
\textbf{Method} & \textbf{\nSHD}$\downarrow$ & \textbf{AUPRC}$\uparrow$  & \textbf{\nSHD}$\downarrow$ & \textbf{AUPRC}$\uparrow$  \\ \hline
\dynotears        & 0.675(0.109)            & 0.224(0.045) & 0.625            & 0.590   \\
\pcmci            & 0.618(0.095)            & 0.244(0.039) & 0.500            & 0.306    \\
\varlingam        & 0.500(0.029)            & 0.386(0.000) & 0.500            & 0.441  \\ 
\ourmethod (RBF) & \textbf{0.250(0.000)}   & \textbf{0.791(0.000)} & \textbf{0.250}            & \textbf{0.791}  \\
\ourmethod (Poly) & \textbf{0.250(0.000)}   & \textbf{0.791(0.000)}& 0.375           & 0.687   \\\hline
\end{tabular}
\caption{\nSHD (Lower is better) and \aupr (Higher is better) for simulations on the synthetic Double-Mass spring system and for the Real Double-Linear Spring system. The ground truth contains four variables and six edges. \ourmethod ranks highest for both metrics when using the RBF kernel. An empty graph results on the real version when using a polynomial kernel.}\label{tab:appmass}
\end{table*}

\begin{table*}[t!]
\centering
\begin{tabular}{ccc}
\hline
\textbf{Method} & \textbf{\nSHD}$\downarrow$ & \textbf{AUPRC}$\uparrow$  \\ \hline
\dynotears        & 0.314(0.163)            & 0.336(0.187) \\
\pcmci            & 0.337(0.046)    &  0.219(0.054)\\
\varlingam        & 0.507(0.099)    & 0.299(0.066) \\ 
\ourmethod (RBF) & 0.170(0.018)   & 0.553(0.121) \\
\ourmethod (Poly) & \textbf{0.166(0.021)}   & \textbf{0.592(0.130)} \\\hline
\end{tabular}
\caption{\nSHD (Lower is better) and \aupr (Higher is better) for the Rössler Oscillator. The ground truth contains 10 variables and Y edges. \ourmethod ranks highest for both metrics.}\label{tab:rossler}
\end{table*}

\section{Computational Analysis}\label{appendix:complexity}

\paragraph{Computation Complexity} The edge scoring phase performs a $\mathcal{O}(D^2)$ pairwise comparisons each involving GP-regression and priority queue updates whose complexity are $\mathcal{O}(N^3)$ resp. $\mathcal{O}(\log D)$, resulting in complexity of $\mathcal{O}(N^3D^2\log D )$. Both forward and backward search phases consider each present edge once and subsequently update at most $\mathcal{O}(D)$ edge scores, giving us a loose upper bound of $O(N^3D^3\log D)$, which is the overall complexity, by parallelizing this computation over edges, we can bring this down to $O(N^3D^2)$ in the best case. 

This is better than worst-case complexity of \pcmci with Kernelized independence test, which is in $\mathcal{O}(N^3 2^D)$ in the worst-case. The complexity for \varlingam is in $\mathcal{O}(D^3)$ because it assumes a parametric form of function between variables. We, nevertheless, see from our evaluation that while faster in practice, it achieves a sub-par performance. For \dynotears there is no explicit Big‑O bound given in the original work but we note that evaluating the acyclicity constraint requires computing the matrix exponential of a $D \times D$ adjacency matrix, which incurs a cost of $O(D^3)$ at the very least. 

\paragraph{Runtime Analysis} In Table n:\ref{tab:runtime}, we provide runtime measurements on the Erdos-Renyi experiment. The main driver for its scalability is the chosen model class. Indeed, in our case the runtime is dominated by the time needed to train the GPs. Stronger scalability can be achieved by substituting GPs with a more lightweight model class.

\begin{table*}[t!]
\centering
\begin{tabular}{cccc}
\hline 
 & \multicolumn{3}{c}{Runtime$\downarrow$} \\\hline
\textbf{Method} & \textbf{ER-5} & \textbf{ER-10} & \textbf{ER-15}   \\ \hline
\dynotears        & 0.28(0.267) & 1.75(1.66) & 3.86(3.49)  \\
\pcmci            & 12.38(11.014) & 309.97(1513.29) & 84.25(40.05)  \\
\varlingam        & 0.12(0.031) & 0.54(0.17) & 1.48(0.50)  \\ \hline
\ourmethod (AB1/RBF) & 950.64(1834.42) & 2025.34(1389.26) & 4234.75(3125.90)  \\
\ourmethod (AB2/RBF) & 1292.11(1236.24) & 3653.92(2276.15) & 7448.01(3365.10)  \\
\ourmethod (AB3/RBF) & 2241.12(2437.72) & 6920.47(3559.90) & 9617.29(3735.26) \\ \hline
\ourmethod (AB1/Poly) & 577.25(2082.06) & 484.63(67.17) & 940.66(149.99) \\
\ourmethod (AB2/Poly) & 173.60(22.14) & 790.25(162.29) & 1564.18(238.23) \\
\ourmethod (AB3/Poly) & 280.244(43.85) & 1178.49(199.63) & 2571.59(396.89) \\ \hline
\end{tabular}
\caption{[Runtime for ER graphs of increasing size on regular timelines]: The runtime for \ourmethod~ is highly dependent on the chosen model class. For our GP models, their training dominates the rest of the greedy-DAG search. Improved scalability can be achieved upon choosing a more lightweight model class.}\label{tab:runtime}
\end{table*}

\section{Example of Scoring}\label{sec:scoring_example}
Let us assume a simple scenario with variables $A, B$ and $C$. Being a score-based method, \ourmethod will search for the combination of parents minimizing our designed score. Therefore, \ourmethod needs to decide between the sets $\{B\}$, $\{C\}$, or $\{B, C\}$, and will do so by searching (greedily) the one with with lowest score. In this section, we present an operative and step-by-step example on how to score the GP when the candidate parents are $B$ and $C$.
\subsection{Scoring $B \rightarrow A \leftarrow C$:} We want to show how our score will be computed when the parent set of $A$ is $\{B, C\}$. We will go through how each term of the score is computed

First, we fit a local model $M_A$, and later we score that model. As a model class, we use the GPs in \citep{ensinger2024continuous}
to learn the continuous-dynamics associated to $A$ as a function of $B$ and $C$, which is $A = F_A (\{B, C\})$. 
After training the GP, the scoring procedure can start. What is necessary is to encode the model's parameters (\textit{function cost}), its causal parents (\textit{structure cost}), and the \textit{data-given-model cost}. Function cost and structure cost are the two parts of Eq.\ref{eq:model-node}, which we recall below
\begin{equation}
L(M_i) = \underbrace{L_{\mathbb{N}}(||\pa_i||) + ||\pa_i||\log D}_{\text{Structure Cost}} + \underbrace{L_{F}(F_i)}_{\text{Function Cost}} \; , 
\end{equation}

\paragraph{Computing the Function Cost} The function cost is computed by encoding the length scale $\alpha_i$, variance $\beta_i$, and eigenvalues in $\Lambda_i$ using $L_p$. Each one of those is encoded by using $L_p$ defined in Eq.\ref{eq:encode-params}. In practice, the function cost $L_{F}(F_A)$ is obtained by summing all the entries of the encoded length-scale vector, the encoded noise variance, and the encoded eigenvalues in $\Lambda_i$. 
\paragraph{Computing the Structure Cost} The structure cost encodes the chosen causal parents, and it is composed by two terms. The first one is just an optimal encoding for integers ~\citep{rissanen:83:integers} for the number of causal parents, that in our case are $B$ and $C$, implying that $||\pa_A|| = 2$. Similarly, we also have another term $||\pa_A||\log D = 2 \cdot \log D$.
Therefore, the structure cost is $L_{\mathbb{N}}(2) + 2\log D$.
\paragraph{Computing Data-given-Model Cost} We use the GP for $F_A$ to predict the trajectory $\tilde{X}_A (t)$ over the timeline $\{t_1,..., t_{N}\}$. We then need to compute an empirical estimate of the residual noise variance:
\begin{equation}
\hat{\sigma}_{A}^2 = \sqrt{\frac{1}{N} \sum_{i = 1}^N \Big(\tilde{X}_A^{(t_i)} - X_A^{(t_i)}\Big)^2 }
\end{equation}
where $X_A^{(t_i)}$ refers to the ground truth trajectory for $A$. The resulting quantity is then used to encode $\nu_{A}$ using Eq.\eqref{eq:residuals} and resulting in $L(\nu_{A})$. 
\paragraph{Summing All the Terms} Finally, all the above-mentioned terms have to be summed together, yielding
\begin{equation}
L(M_A) = L_{\mathbb{N}}(2) + 2\log D + L_{F}(F_A) + L(\nu_{A})
\end{equation}
\subsection{Scoring with no Parents}
During the initial edge scoring phase (See algorithm \ref{alg:edgescoring}), scores need to be computed also for the case in which there are no parents. In this case, we just encode the mean value of $X_A^{(t)}$ over $t$ which we indicate as $\mathbb{E}_{t}[X_A^{(t)}]$.

\paragraph{Function Cost} This time there are no parents or self-loops. Therefore, instead of scoring a trained model, we just encode the mean value of $X_A^{(t)}$, i.e.\, we use Eq.\ref{eq:encode-params} to encode $\mathbb{E}_{t}[X_A^{(t)}]$.

\paragraph{Computing Data-given-Model cost} Our estimate of the residual noise variance is consequently computed between $X_A^{(t)}$ and its mean. Therefore, we have
\begin{equation}
\hat{\sigma}_{A}^2 = \sqrt{\frac{1}{N} \sum_{i = 1}^N (\mathbb{E}_{t}[X_A^{(t)}] - X_A^{(t_i)})^2 }
\end{equation}

\section{Algorithms Pseudo-code}\label{sec:pseudocode}
\begin{algorithm}[]
\caption{\ourmethod for timeseries causal discovery}\label{alg:mdl_gp}\label{alg:cadyt}
\textbf{Input}: Timeseries Trajectory \trajectory over $\mathit{\variables = \{X_1,...,X_D\}}$\\
\textbf{Output}: Causal graph $\tgraph$~over \variables \\
\begin{algorithmic}[1] 
\STATE $\mathit{Q} \leftarrow \textsc{EdgeScoring}\mathit{(\trajectory)}$\\
\STATE $\tgraph \leftarrow \textsc{ForwardSearch}\mathit{(Q,\trajectory)}$ \\
\STATE $\tgraph \leftarrow \textsc{BackwardSearch}\mathit{(\tgraph,\trajectory)}$ \\
\STATE $\mathit{return~\tgraph}$
\end{algorithmic}
\end{algorithm}

For ease of reading, we again describe our score-based method \textbf{Ca}usal Discovery for \textbf{Dy}namic \textbf{T}imeseries (\ourmethod\!\!) for discovering causal graphs of multivariate continuous-valued dynamical systems. We incorporate our proposed score into a common three-step search procedure~\citep{mian:21:globe,mameche:23:linc} namely, edge scoring, forward and backward search, as shown in Alg.~\ref{alg:cadyt}. This is the next best alternative to exhaustive search for our case. The well known Greedy Equivalence Search~\cite{chickering:02:ges} is not built for timeseries and methods for topological search~\cite{wang:igsp,xu:25:topic} do not directly apply to cyclic systems. We provide our code in the supplementary material.

We start with the edge-ranking phase that computes the gain for all pairwise causal connections. The gain $\gain_{ij}$, between each pair $X_i$ and $X_j$, is given by
\begin{equation}
    \gain_{ij} = L(\trajectory,M) - L(\trajectory,\Gright) 
\end{equation}
where \Gright~ implies model $M$ with edge $X_i \to X_j$ included. Intuitively, the higher the $\gain_{ij}$, the more confident we are that this is the correct causal edge. The edge scoring phase returns a priority queue of tuples $(\gain_{ij},(X_i,X_j))$ ordered by decreasing gain.

The forward search iteratively adds the highest-scoring edge from the priority queue. After adding an edge \(X_i \to X_j\), all incoming edges to \(X_j\) are re-evaluated using Eq.~\eqref{eq:delta1} and updated in the queue. Before inclusion, each edge is tested for statistical significance using the no-hypercompression inequality~\citep{grunwald:07:mdl}. The search terminates when no further additions improve the score. The subsequent backward phase prunes redundant edges by removing any whose deletion increases \(L(\trajectory, M)\), continuing until no such edges remain. We then return the final graph. 

\ourmethod has overall computational complexity of $\mathcal{O}(N^3D^3\log D)$ where $N^3$ derives from our choice of non-parametric regression functions (GPs) for $L_F$. In appendix~\ref{appendix:complexity}, we give a detailed derivation on the computational complexity and show how it is at least on-par with existing methods. \ourmethod, moreover, can be inherently parallelized, and we implement it as such, resulting in a fast runtime. 

\begin{algorithm}[t!]
\caption{\textsc{EdgeScoring}}\label{alg:edgescoring}
\textbf{Input}: Timeseries Trajectory \trajectory over $\mathit{\variables = \{X_1,...,X_D\}}$, local scoring function $S_{local}(\trajectory,i, Pa_i)$\\
\textbf{Output}: Score matrix $\mathit{Q}$~over \variables \\
\begin{algorithmic}[1] 
\WHILE{$i,j = 1, \dots, D$}
    \STATE $\mathit{Q}_{ij} \leftarrow S_{local}(\trajectory,~j,~i)$
\ENDWHILE
\STATE $\mathit{return~\mathit{Q}}$
\end{algorithmic}
\end{algorithm}

\begin{algorithm}[t!]
\caption{\textsc{ForwardSearch}}\label{alg:forwardsearch}
\textbf{Input}: Timeseries Trajectory \trajectory over $\mathit{\variables = \{X_1,...,X_D\}}$, global scoring function $S(\tgraph, \trajectory)$, scalar threshold $\alpha$, edge scoring matrix $Q$\\
\textbf{Output}: Causal graph $\tgraph$~over \variables \\
\begin{algorithmic}[1] 
\STATE $Done \gets \text{False}$ \\
\STATE $CandidateEdges \gets \{\}$
\FOR{\textbf{each} $i,j = 1, \dots, D$}
\IF{$ Q_{ij} >\alpha$}
\STATE $CandidateEdges \gets CandidateEdges \cup (i \to j)$
\ENDIF
\ENDFOR
\STATE $CandidateEdges \gets DecreasingOrder(CandidateEdges)$
\WHILE{$\neg Done$}
\STATE $\Delta S_{best} \gets - \infty$
\STATE $E_{best} \gets \text{None}$
\FOR{\textbf{each} $(X\rightarrow Y) \in CandidateEdges$}
\IF {$E \notin G_{current}$}
\STATE $\tgraph_{candidate} \gets \tgraph_{current} \cup (X\rightarrow Y)$
\STATE $S_{candidate} \gets S(\tgraph_{candidate}, \trajectory)$
\STATE $S_{current} \gets S(\tgraph_{current}, \trajectory)$
\STATE $\Delta S \gets S_{candidate} - S_{current}$
\IF{$ \Delta S > \Delta S_{best}$}
\STATE $\Delta S_{best} \gets \Delta S$
\STATE $E_{best} \gets (X\rightarrow Y)$
\ENDIF
\ENDIF
\ENDFOR
\IF{$ \Delta S_{best} >\alpha$}
\STATE $\tgraph_{current} \gets \tgraph_{current} \setminus E_{best}$
\ELSE
\STATE $Done \gets True$
\ENDIF

\ENDWHILE
\STATE $\mathit{return~\tgraph}$
\end{algorithmic}
\end{algorithm}

\begin{algorithm}[]
\caption{\textsc{BackwardSearch}}\label{alg:backwardsearch}
\textbf{Input}: Timeseries Trajectory \trajectory over $\mathit{\variables = \{X_1,...,X_D\}}$, scoring function S, scalar threshold $\alpha$\\
\textbf{Output}: Causal graph $\tgraph$~over \variables \\
\begin{algorithmic}[1] 
\STATE $Done \gets \text{False}$ \\
\WHILE{$\neg Done$}
\STATE $Edges \gets \text{All edges in }\tgraph_{current}$
\STATE $\Delta S_{best} \gets - \infty$
\STATE $E_{best} \gets \text{None}$
\FOR{\textbf{each} $(X\rightarrow Y) \in Edges$}
\STATE $\tgraph_{candidate} \gets \tgraph_{current} \setminus (X\rightarrow Y)$
\STATE $S_{candidate} \gets S(\tgraph_{candidate}, \trajectory)$
\STATE $S_{current} \gets S(\tgraph_{current}, \trajectory)$
\STATE $\Delta S \gets S_{candidate} - S_{current}$
\IF{$ \Delta S > \Delta S_{best}$}
\STATE $\Delta S_{best} \gets \Delta S$
\STATE $E_{best} \gets (X\rightarrow Y)$
\ENDIF
\ENDFOR

\IF{$ \Delta S_{best} >\alpha$}
\STATE $\tgraph_{current} \gets \tgraph_{current} \setminus E_{best}$
\ELSE
\STATE $Done \gets True$
\ENDIF

\ENDWHILE
\STATE $\mathit{return~\tgraph}$
\end{algorithmic}
\end{algorithm}

\end{document}